\documentclass[english,twocolumn,transaction]{IEEEtran}
\usepackage[T1]{fontenc}
\usepackage{arydshln}
\usepackage{float}
\usepackage{amsmath}
\usepackage{amsmath,lipsum}
\usepackage{cuted}
\usepackage{amsthm}
\usepackage{mathrsfs}
\usepackage{amssymb}
\usepackage{stackrel}
\usepackage{graphicx}
\usepackage{setspace}
\usepackage{url}
\usepackage[ruled,linesnumbered]{algorithm2e}
\usepackage{booktabs}
\usepackage{textcomp,booktabs}
\usepackage[usenames,dvipsnames]{color}
\usepackage{colortbl}
\usepackage{color}
\usepackage{cite}
\usepackage{stfloats}
\usepackage{multirow}
\usepackage{subfigure}

\newcommand{\tabincell}[2]{\begin{tabular}{@{}#1@{}}#2\end{tabular}}
\makeatletter
\newcommand{\removelatexerror}{\let\@latex@error\@gobble}

\newcommand\figcaption{\def\@captype{figure}\caption}
\newcommand\tabcaption{\def\@captype{table}\caption}

\theoremstyle{plain}

\theoremstyle{definition}

\theoremstyle{plain}

\theoremstyle{plain}

\IEEEoverridecommandlockouts
\usepackage{ifpdf}
\usepackage{cite}
\ifCLASSOPTIONcompsoc
\usepackage[caption=false,font=normalsize,labelfont=sf,textfont=sf]{subfig}
\else
\usepackage[caption=false,font=footnotesize]{subfig}
\fi
\usepackage{amsmath}
\@ifundefined{showcaptionsetup}{}{%
 \PassOptionsToPackage{caption=false}{subfig}}
\usepackage{subfig}
\makeatother

\usepackage{babel}

\newtheorem{crit}{Criterion}

\renewcommand\figurename{Fig.}

\definecolor{mygray}{gray}{.9}
\definecolor{mygray1}{gray}{1.}
\definecolor{mypink}{rgb}{.99,.91,.95}
\definecolor{mycyan}{cmyk}{.3,0,0,0}

\begin{document}
\title{Multi-Agent Deep Reinforcement Learning for Cost- and Delay-Sensitive Virtual Network Function Placement and Routing}

\author{Shaoyang Wang, Chau Yuen,~\IEEEmembership{Fellow,~IEEE}, Wei Ni,~\emph{Senior Member, IEEE},\\ Guan Yong Liang,~\emph{Senior Member, IEEE}, and Tiejun Lv,~\emph{Senior Member, IEEE}

\thanks{Manuscript received January 5, 2022; revised May 2, 2022; accepted June 16, 2022. This work was supported by the China Scholarship Council, and part of the work was done during the visit to SUTD in 2021. \emph{(Corresponding author: Tiejun Lv)}.

S. Wang and T. Lv are with the School of Information and Communication Engineering, Beijing University of Posts and Telecommunications, Beijing 100876, China (e-mail: \{shaoyangwang, lvtiejun\}@bupt.edu.cn).

C. Yuen is with the Engineering Product Development, SUTD, 8 Somapah Road, Singapore 487372 (e-mail: yuenchau@sutd.edu.sg).

W. Ni is with the Commonwealth Scientific and Industrial Research Organisation (CSIRO), Sydney 2122, Australia (e-mail: Wei.Ni@data61.csiro.au).

Y. Guan is with the School of Electrical and Electronic Engineering, NTU, Singapore 639798 (e-mail: eylguan@ntu.edu.sg).

}}
\maketitle
\begin{abstract}
This paper proposes an effective and novel multi-agent deep reinforcement learning (MADRL)-based method for solving the joint virtual network function (VNF) placement and routing (P\&R), where multiple service requests with differentiated demands are delivered at the same time. The differentiated demands of the service requests are reflected by their delay- and cost-sensitive factors. We first construct a VNF P\&R problem to jointly minimize a weighted sum of service delay and resource consumption cost, which is NP-complete. Then, the joint VNF P\&R problem is decoupled into two iterative subtasks: placement subtask and routing subtask. Each subtask consists of multiple concurrent parallel sequential decision processes. By invoking the deep deterministic policy gradient method and multi-agent technique, an MADRL-P\&R framework is designed to perform the two subtasks. The new \emph{joint reward and internal rewards} mechanism is proposed to match the goals and constraints of the placement and routing subtasks. We also propose the parameter migration-based model-retraining method to deal with changing network topologies. Corroborated by experiments, the proposed MADRL-P\&R framework is superior to its alternatives in terms of service cost and delay, and offers higher flexibility for personalized service demands. The parameter migration-based model-retraining method can efficiently accelerate convergence under moderate network topology changes.
\end{abstract}
\begin{IEEEkeywords}
Placement and routing, multi-agent deep reinforcement learning, virtual network functions
\end{IEEEkeywords}
\section{Introduction}
\renewcommand\figurename{Fig.}
Network function virtualization (NFV) applies virtualization (or softwarization) technologies to decouple software instances from hardware platforms \cite{7534741}. Network function that relied on proprietary hardware in the past can be deployed on general-purpose hardware (e.g., high volume server, HVS) in the form of software. With many advantages such as flexibility and scalability, NFV has been regarded as the future network service paradigm \cite{9169857}. In NFV, the requested network service is composed of an ordered set of virtual network functions (VNFs). These VNFs are run in a predefined order to form an ordered list of network functions, i.e., service function chain (SFC). The efficiency of NFV-based platforms heavily depends on how effectively network resources are allocated for the requested SFCs, that is, these VNFs placement and routing.

The VNF placement-and-routing (P\&R) problem has received significant attention. The VNF placement and/or routing problems are typically formulated as integer linear programming (ILP) or mixed ILP (MILP) problems. The optimization of VNF P\&R with different objectives or demands, such as delay minimization, cost minimization, profit maximization, and load balancing, is computationally prohibitive due to the factorial nature of the VNF routing \cite{rost2018charting}. While branch-and-bound can provide exact solutions for small-scale system models, heuristics and meta-heuristics \cite{8556457} are often applied under relatively large-scale problem settings.

Several heuristics have been developed for different optimization objectives and application scenarios. Alhussein \emph{et al.} \cite{8648029} modeled the VNF P\&R as an  MILP problem that minimizes the service cost, and proposed a low-complexity heuristic algorithms. Tajiki \emph{et al.} \cite{8480442} designed heuristic algorithms for solving the joint QoS-aware and energy efficient VNF P\&R problem. The pivotal idea of these heuristic algorithms is to find the closest server to the first VNF in SFC. Liu \emph{et al.} \cite{9468711} studied a dynamic VNF P\&R problem to minimize the resource consumption. After formulating this optimization problem as an ILP problem, the authors constructed a heuristic optimization structure: VNF-splited multi-stage edge-weight graph. Varasteh \emph{et al.} \cite{9340363} formulated the joint VNF P\&R problem about power and delay as an ILP problem, and proposed an online heuristic structure to solve this optimization problem. In \cite{7859379}, the authors studied a heuristic method combined with genetic algorithm to shorten the computation time of VNF placement. These heuristics can speed up the solution and can hardly guarantee high quality solutions \cite{mostafavi2021quality}. In addition to heuristics, matching theory \cite{7859379}, game theory \cite{7421947,8314673}, queuing theory \cite{8611305,8485943}, and column generation model \cite{7938396} have also been applied to solve the VNF P\&R problem or its variants. These algorithms typically suffer from exponential running time \cite{kaur2020comprehensive}.

How to determine effective placement and routing for VNFs remains as a challenging and critical problem in a NFV-based network. Moreover, existing works have generally overlooked personalized differences in service requests of different users. This paper addresses a new joint VNF placement and routing problem with more complex optimization objectives and more abundant constraints, where 1) the resource consumption cost and service delay are modeled more comprehensively, and then are simultaneously minimized, 2) different users have different demands of delay and cost, parameterized by new delay- and cost-sensitive factors, and 3) multiple service requests are deployed in parallel, instead of being deployed serially. The problem is more challenging, and a new and more effective solution is expected.

Recently, machine learning has been increasingly applied to NFV-based networks. Li \emph{et al.} \cite{8501525} studied the VNF chaining in an optical network by employing a deep learning (DL)-based prediction technique. Emu \emph{et al.} \cite{9492772} used convolutional neural network (CNN) to minimize communication costs of VNF deployment. Pei \emph{et al.} \cite{9062539} divided the VNF selection and chaining problem into two stages: first VNF selection, and then VNF chaining, and developed a  DL-based two-phase algorithm comprising VNF selection and VNF chaining networks. The above DL-based method is essentially supervised learning, and would rely heavily on effective acquisition of meaningful datasets for effective training of the DL models.

Different from DL, reinforcement learning (RL) can explore solutions on its own \cite{8714026}. Li \emph{et al.} \cite{9075271} used Q-learning to  solve a VNF scheduling problem with minimum makespan of all services. Bunyakitanon \emph{et al.} \cite{9070195} also presented a Q-learning-based module which automates VNF placement. Dalgkitsis \emph{et al.} \cite{9322512} leveraged the deep deterministic policy gradient algorithm to implement dynamic resource-aware VNF placement. Akbari \emph{et al.} \cite{9450026} studied a VNF placement and scheduling problem in an industrial internet-of-things network, and applied an actor-critic RL to jointly minimize the VNF cost and the age of information.

Existing DRL-based methods typically work in a fixed network topology. When the network topology changes (such as the increase or decrease of nodes), it is usually necessary to build a new neural network and retraining, due to the number of neurons in the input and output layers is related to the network topology. To the best of our knowledge, no existing study has addressed the adaptivity of DRL in a dynamic network topology. In addition, we consider that multiple service requests with differentiated demands are deployed at the same time, which incurs a huge action space. The above existing DRL-based methods are invalid in this problem.

In this paper, we propose, for the first time, a multi-agent deep reinforcement learning (MADRL) approach to VNF P\&R problem, where the service delay and resource consumption cost of VNF P\&R are  jointly minimized. In particular, our proposed framework can better support the dynamic changes of network topology. The key contributions of this paper include:
\begin{itemize}
\item A new VNF P\&R problem is formulated to jointly minimize resource consumption cost and service delay of multiple concurrent service requests. The demand difference of each request is parameterized by new delay- and cost-sensitive factors. Next, this optimization problem is decomposed into two DRL subtasks: placement and routing subtasks. Each subtask is transformed into multiple concurrent parallel sequential decision processes.
\item A novel MADRL-based P\&R (MADRL-P\&R) framework is proposed for these two DRL subtasks. The MADRL-P\&R framework consists of two multi-agent modules: a placement module responsible for VNF placement of multiple service requests and a routing module responsible for VNF routing of multiple service requests. The deep deterministic policy gradient (DDPG) algorithm is applied to the implementation of each agent in the two modules. The \emph{joint reward and internal rewards} mechanism is also designed to achieve the objective and constraints of these two subtasks.
\item A novel parameter migration-based model-retraining method is proposed to deal with changes in the network topology. In order to eliminate the dependence of the neural network input layer structure on the number of NFVs required by different service requests, the structure of the input information is meticulously designed to facilitate the migration of the network.
\end{itemize}

Experiments on a ``COST266'' network from the ``SNDlib'' \cite{orlowski2010sndlib} verify that the proposed MADRL-P\&R framework provides near-optimal results and is superior to the state of the art in terms of service cost and delay. The framework can achieve a more reasonable resource allocation to increase the number of service requests deployed concurrently, thereby improving network throughput. Compared with its existing alternatives, the proposed MADRL-P\&R method offers excellent flexibility to meet the differentiated demands of the users.

The rest of this paper is as follows. The network is modeled in Section \uppercase\expandafter{\romannumeral2}, and the VNF P\&R optimization problem is formulated in Section \uppercase\expandafter{\romannumeral3}. In Section \uppercase\expandafter{\romannumeral4}, the optimization problem can be transformed into the DRL tasks. The MADRL-P\&R framework is designed in Section \uppercase\expandafter{\romannumeral5}. Experimental results are assessed in Section \uppercase\expandafter{\romannumeral6}, and in Section \uppercase\expandafter{\romannumeral7}, this paper is concluded. Notations used are collected in Table \uppercase\expandafter{\romannumeral1}.
\begin{table*}[hbp]
\setstretch{0.5}
   \scriptsize
	\centering{}
	\textbf{Table \uppercase\expandafter{\romannumeral1}}~~The symbol and description used in Section \uppercase\expandafter{\romannumeral2}--Section \uppercase\expandafter{\romannumeral4}.\\
\setlength{\tabcolsep}{0.4mm}{
\begin{tabular}{|c|p{160pt}|c|p{220pt}|}
\hline
\textbf{Symbol}&\textbf{Description} &\textbf{Symbol} &\textbf{Description} \\
\hline
\multicolumn{2}{|>{\columncolor{mygray}}c|}{\textbf{Section \uppercase\expandafter{\romannumeral2}}}&$N_{\mathrm{hop},i}$&The total hop count of the routing chain of $S_i$\\
$\mathcal{G}$& The network graph model of the SN&\multicolumn{2}{>{\columncolor{mygray}}c|}{\textbf{Section \uppercase\expandafter{\romannumeral3}}}\\
$\mathcal{V}$&The node set of $\mathcal{G}$&$\mathbf{H}_{i}$ & The data flow size matrix of $S_i$\\
$\mathcal{E}$&The link set of $\mathcal{G}$&$\varGamma(P_{i,(j,n)})$&The VNF category of the $j$-th VNF in $S_i$ placed on node $n$\\
$N$&The number of nodes in $\mathcal{G}$&$c_{n}^{\mathrm{c}}$&The cost per unit of computing resource consumed at node $n$\\
$L$&The number of links in $\mathcal{G}$&$c_{n}^{\mathrm{m}}$&The cost per unit of memory resource consumed at node $n$\\
$\left(u,v\right)$&The directed flow from node $u$ to node $v$&$c_{(u,v)}^{\mathrm{B}}$&The cost per unit of bandwidth resources consumed at link $(u,v)$\\
$\mathbf{A}$&The adjacency matrix of $\mathcal{G}$&$d_{(u,v)}^{\mathrm{B}}$ & The delay of link $(u,v)$ per unit service rate\\
$\mathcal{F}$&The VNF set requested by service requests&$d^{\mathrm{inv}}$ & The invariant processing delay of each node\\
$N_{\mathrm{f}}^{\mathrm{total}}$ & The number of VNF categories&$d_{\varGamma(P_{i,(j,n)})}^{\mathrm{dyn}}$ & The dynamic processing delay of the $\varGamma(P_{i,(j,n)})$-th category VNF\\
$F_k$ & The $k$-th category VNF&$\varphi_{i}^{\mathrm{c}}$ & The cost-aware factor of $S_i$\\
$C_n$&The computing capacity resources of node $n$&$\varphi_{i}^{\mathrm{d}}$& The demand-aware factor of $S_i$\\
$M_n$&The memory capacity of node $n$&$\psi^{\mathrm{c}}$ & The coefficient to eliminate order of magnitude differences\\
$f_{k}^{\mathrm{m}}$ & The memory size required by the $k$-th category VNF&$\psi^{\mathrm{d}}$ & The coefficient to eliminate order of magnitude differences\\
$B_{\left(u,v\right)}$&The bandwidth capacity of link $\left(u,v\right)$&\multicolumn{2}{>{\columncolor{mygray}}c|}{\textbf{Section \uppercase\expandafter{\romannumeral4}}}\\
$f_{k}^{\mathrm{c}}$ & The computing capacity required by the $k$-th category VNF&$\gamma$ & The reward discount factor\\
$\eta_{k}$& The computing resource consumed per unit service rate of the $k$-th category VNF&$\mathbf{a}_{i,t}^{\mathrm{p}}$ & The placement action of the $i$-th agent at decision step $t$\\
$r^{\mathrm{s}}$ & The service rate&$\mathbf{a}_{i,t}^{\mathrm{r}}$ & The routing action of the $i$-th agent at decision step $t$\\
$S_{i}$&The $i$-th service request&$s_{i,t}^{\mathrm{p}}$ & The placement state of the $i$-th agent at decision step $t$\\
$M$ &The number of service requests&$s_{i,t}^{\mathrm{r}}$ & The routing state of the $i$-th agent at decision step $t$\\
$s_i$&Source node of $S_i$&$\mathbb{\mathring{F}}_{i,t}^{\mathrm{re}}$ & VNFs in $S_i$ that have not yet been placed at decision step $t$\\
$t_i$&Destination node of $S_i$&$\mathbf{C}_{t}^{\mathrm{re}}$ & The remaining computing capacity available at decision step $t$\\
$\mathbb{F}_{i}$&The SFC of $S_i$&$\mathbf{M}_{t}^{\mathrm{re}}$ & The remaining memory size available at decision step $t$\\
$r^{\mathrm{s}}_i$ & The service rate of $S_i$&$\mathbf{B}_{t}^{\mathrm{re}}$ & The remaining bandwidth of link $(u,v)$ at decision step $t$\\
$\mathscr{F}_{i,j}$&The $j$-th VNF required by $S_{i}$&$\mathbf{R}_{i,t}^{\mathrm{cu}}$ & The current routing matrix\\
$N_{\mathrm{f},i}$ & The number of VNFs requested by $S_i$&$r_{i,t}^{\mathrm{p,int}}$ & The internal reward of the $i$-th placement agent at decision step $t$\\
$\mathbf{P}_{i}$ & The placement matrix of $S_i$&$r_{i,t}^{\mathrm{r,int}}$ & The internal reward of the $i$-th routing agent at decision step $t$\\
$\mathbf{R}_{i}$ & The routing matrix of $S_i$&$r_{i,t}^{\mathrm{p,jo}}$ & The joint reward of the $i$-th placement agent at decision step $t$\\
$\mathbf{Q}_{i}$ & The hop matrix of $S_i$&$r_{i,t}^{\mathrm{r,jo}}$ & The joint reward of the $i$-th routing agent at decision step $t$\\
\hline
\end{tabular}}
\end{table*}

\section{System Model}
\subsection{Substrate Network and NFV Infrastructure}
Define an undirected graph $\mathcal{G}\!=\!(\mathcal{V},\mathcal{E})$ to describe a substrate network (SN), including $N$ nodes and $L$ links. Let $\mathcal{V}$ and $\mathcal{E}$ denote the sets of nodes and links, then $|\mathcal{V}|\!=\!N$ and $|\mathcal{E}|\!=\!L$. Define $(u,v)$ represent the directed flow from node $u$ to node $v$. The topology of $\mathcal{G}$ is characterized by its adjacency matrix $\mathbf{A}\!=\![A_{(u,v)}]_{N\times N}$. If $(u,v)\in\mathcal{E}$, $A_{(u,v)}\!=\!1$; otherwise, $A_{(u,v)}\!=\!0$.

All VNFs requested by services are collected by $\mathcal{F}=\{ F_{1},F_{2},\ldots,F_{N_{\mathrm{f}}^{\mathrm{total}}}\}$, where $N_{\mathrm{f}}^{\mathrm{total}}$ is the number of VNF categories. The NFV infrastructure (NFVI) provides different network resources to host the VNFs. We use $C_{n}$ to denote the computing capacity of node $n$ and $M_{n}$ to denote the memory capacity of node $n$. The resources required for the placement of the $k$-th category VNF, $F_{k}$ ($k\in\left\{ 1,\ldots,N_{\mathrm{f}}^{\mathrm{total}}\right\} $), can be described as $\left\{ f_{k}^{\mathrm{m}},f_{k}^{\mathrm{c}}\right\} $, where $f_{k}^{\mathrm{m}}$ is the memory size required by the $k$-th category VNF and $f_{k}^{\mathrm{c}}$ is the computing capacity required by the VNF. Since the required computing power depends on the amount of data to be processed, $f_{k}^{\mathrm{c}}$ is given by $f_{k}^{\mathrm{c}}=\eta_{k}r^{\mathrm{s}}$, where $r^{\mathrm{s}}$ is the service rate and $\eta_{k}$ is the computing resource consumed per unit service rate of the $k$-th category VNF. In addition, for each link, we use $B_{(u,v)}$ to denote the bandwidth capacity of link $\left(u,v\right)$, and $B_{(u,v)}\!=\!B_{(v,u)}$.
\subsection{Service Function Chain Request}
Let $\mathcal{S}\!=\!\{ S_{1},\ldots,S_{i},\ldots,S_{M}\}$ denote service requests of the $M$ users, where $S_{i}$ is the service request of the $i$-th user. $S_{i}$ is described by $\{ s_{i},t_{i},\mathbb{F}_{i},r_{i}^{\mathrm{s}}\}$, where $s_{i}$ is the source node of $S_{i}$ and $t_{i}$ is the destination node of $S_{i}$; $\mathbb{F}_{i}$ is the SFC of $S_{i}$; and $r^{\mathrm{s}}_{i}$ is the service rate of $S_{i}$. $\mathbb{F}_{i}$ is a set of ordered VNFs and $\mathbb{F}_{i}=\{ \mathscr{F}_{i,1}\ldots\rightarrow\mathscr{F}_{i,j}\rightarrow\ldots \mathscr{F}_{i,N_{\mathrm{f},i}}\} $, where $\mathscr{F}_{i,j}\!\in\!\mathcal{F}$ ($j\in\{ 1,\ldots,N_{\mathrm{f},i}\}$) is the $j$-th VNF requested by $S_{i}$, and $N_{\mathrm{f},i}$ is the number of VNFs requested by $S_{i}$.
\begin{figure*}[htb]
\centering{}\includegraphics[scale=1]{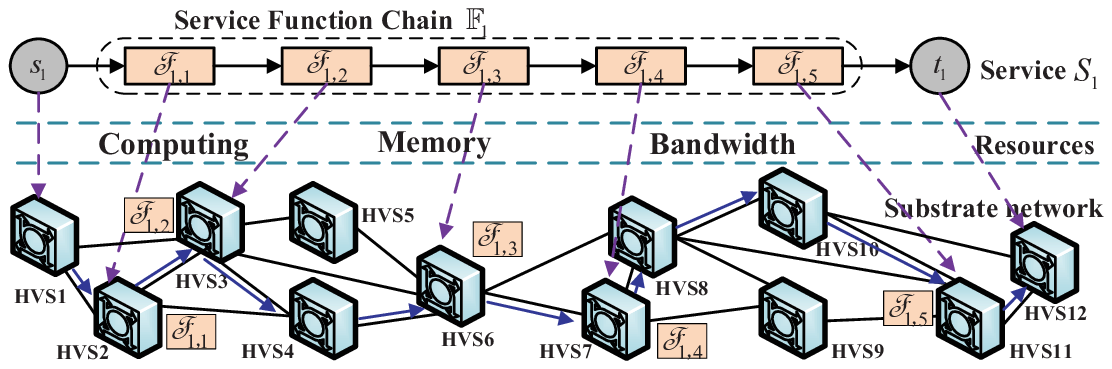}
\caption{The placement and routing of $S_{1}$. $S_{1}=\left\{ s_{1},t_{1},\mathbb{F}_{1},r_{1}^{\mathrm{s}}\right\} $ and $\mathbb{F}_{1}=\left\{ \mathscr{F}_{1,1}\rightarrow\mathscr{F}_{1,2}\rightarrow\mathscr{F}_{1,3}\rightarrow\mathscr{F}_{1,4}\rightarrow\mathscr{F}_{1,5}\right\} $.}\label{system_model}
\end{figure*}

As shown in Fig. \ref{system_model}, this paper considers the joint VNF P\&R. The VNF placement aims to select the suitable nodes to place the VNFs (e.g., $\mathscr{F}_{1,1},\mathscr{F}_{1,2},\mathscr{F}_{1,3},\mathscr{F}_{1,4},\mathscr{F}_{1,5}$) required by service request $S_{1}$, according to the computing/memory resources in NFVIs (e.g., HVSs) and the optimization objective. After placing the VNFs, the VNF routing can be executed, that is, to select the suitable routing path (also known as ``routing chain'') from source $s_{i}$ to destination $t_{i}$, according to the bandwidth resources in NFVIs and the optimization objective. The selected routing path passes all VNFs required in $S_{i}$ in the correct order.

The VNF placement policy of the $i$-th is described by the placement matrix $\mathbf{P}_{i}=[P_{i,(j,n)}]_{N_{\mathrm{f},i}\times N}$. If the $j$-th VNF in $S_{i}$ is placed at the $n$-th node, $P_{i,(j,n)}=1$; otherwise, $P_{i,(j,n)}=0$. A routing chain is described by a \emph{routing matrix} $\mathbf{R}_{i}\!=\![R_{i,(u,v)}]_{N\times N}$ and a \emph{hop matrix} $\mathbf{Q}_{i}\!=\![Q_{i,(u,v)}]_{N\times N}$. The routing matrix $\mathbf{R}_{i}$ describes the direction of the data flow of service request $S_{i}$. If link $(u,v)$ is part of the routing chain for $S_{i}$, $R_{i,(u,v)}=1$; otherwise, $R_{i,(u,v)}=0$. The hop matrix $\mathbf{Q}_{i}$ indicates which hop each link in the routing chain is. If the link $(u,v)$ is the $m$-th hop, $Q_{i,(u,v)}=m$, $\ensuremath{m\in\{1,2,\ldots,N_{\mathrm{hop},i}\}}$, and $N_{\mathrm{hop},i}$ is the total hop count of the routing chain of $S_{i}$. $Q_{i,(u,v)}=0$ meas that the link $(u,v)$ is not part of the routing chain for $S_{i}$. In this paper, the ``typical SFCs'' (acyclic, single-flow, and one-way service requests \cite{8480442}) are considered.
\section{Joint Minimization of Cost and Delay in VNF Placement and Routing}
\subsection{Constraints for VNF Placement and Routing}
\subsubsection{Routing Path Constraint}
Since the typical SFCs are considered, the following constraint needs to be satisfied to ensure the one-wayness:
\begin{align}
\mathrm{C1}:\underset{i\in\left\{ 1,\ldots,M\right\} }{\sum}&\left(\underset{\left(u,v\right)\in\mathcal{E}}{\sum}R_{i,\left(u,v\right)}-\underset{\left(v,u\right)\in\mathcal{E}}{\sum}R_{i,\left(v,u\right)}\right)\nonumber \\
&=\begin{cases}
M, & \mathrm{if}\:u=s_{i};\\
-M, & \mathrm{if}\:u=t_{i};\\
0, & \mathrm{otherwise}.
\end{cases}u,v\in\mathcal{V},\forall i.
\end{align}

Given the unidirectional path, the following constraint needs to be satisfied:
\begin{align}
\mathrm{C2}:R_{i,\left(u,v\right)}+R_{i,\left(v,u\right)}\leq1,\:u,v\in\mathcal{V}.
\end{align}
\begin{crit}
The following constraint should be satisfied for ensuring the acyclic data flow:
\begin{align}
\mathrm{C3}:\sum\nolimits _{\left(u,v\right),\left(v,u\right)\in\mathcal{E}}\left(R_{i,\left(u,v\right)}+R_{i,\left(v,u\right)}\right)\leq2,u,v\in\mathcal{V}.
\end{align}
\end{crit}
\begin{IEEEproof}
We consider three cases: Case i): $u=s_{i}$: If there is no loop, then $\sum\nolimits _{\left(u,v\right),\left(v,u\right)\in\mathcal{E}}(R_{i,(u,v)}+R_{i,(v,u)})=1$\footnote{When $u=s_{i}$, $\sum\nolimits _{\left(u,v\right)\in\mathcal{E}}R_{i,(u,v)}$ represents all service flows flowing out of the source node $s_i$, and $\sum\nolimits _{\left(u,v\right)\in\mathcal{E}}R_{i,(u,v)}$ represents all service flows flowing into the source node $s_i$. The rest is similar.} under the constraint of a single-flow one-way service request. If the number of the loop through $s_{i}$ is $q$ ($q\in\mathbb{N}^{+}$), then $\sum\nolimits _{\left(u,v\right),\left(v,u\right)\in\mathcal{E}}(R_{i,(u,v)}+R_{i,(v,u)})=2q+1$; Case ii): $u=\mathcal{V}-\{ s_{i},t_{i}\}$: By assuming that there is no loop, then $\sum\nolimits _{\left(u,v\right),\left(v,u\right)\in\mathcal{E}}(R_{i,(u,v)}+R_{i,(v,u)})=2$. If $q\geq1$, then $\sum\nolimits _{\left(u,v\right),\left(v,u\right)\in\mathcal{E}}(R_{i,(u,v)}+R_{i,(v,u)})=2q+2$; and Case iii): $u=t_{i}$: Similar to Case i), when the number of loop is $q$, $\sum\nolimits _{\left(u,v\right),\left(v,u\right)\in\mathcal{E}}(R_{i,(u,v)}+R_{i,(v,u)})=2q+1$.
\end{IEEEproof}
\subsubsection{Flow Constraints}
Let $\mathbf{H}_{i}\!=\![H_{i,(u,v)}]_{N\times N}$ denote the data flow size of  $(u,v)$ in the routing chain of $S_{i}$, and $\ensuremath{\mathbf{H}_{i}=r_{i}^{\mathrm{s}}\mathbf{R}}_{i}$. The flow needs to satisfy the flow conservation \cite{6739404}:
\begin{align}
\mathrm{C4}:\underset{\left(u,v\right)\in\mathcal{E}}{\sum}H_{i,\left(u,v\right)}\!-\!\underset{\left(v,u\right)\in\mathcal{E}}{\sum}H_{i,\left(v,u\right)}\!=\!\begin{cases}
r_{i}^{\mathrm{s}}, & \mathrm{if}\:u=s_{i};\\
-r_{i}^{\mathrm{s}}, & \mathrm{if}\:u=t_{i};\\
0, & \mathrm{otherwise}.
\end{cases}
\end{align}
The source and destination nodes also need to satisfy:
\begin{align}
&\mathrm{C5}: \sum\nolimits _{u}R_{i,\left(u,v\right)}=0, \sum\nolimits _{u}Q_{i,\left(u,v\right)}=0, v=s_{i}; \\
&\mathrm{C6}: \sum\nolimits _{v}R_{i,\left(u,v\right)}=1, \sum\nolimits _{v}Q_{i,\left(u,v\right)}=1, u=s_{i}; \\
&\mathrm{C7}: \sum\nolimits _{v}R_{i,\left(u,v\right)}=0, \sum\nolimits _{v}Q_{i,\left(u,v\right)}=0,u=t_{i};\\
&\mathrm{C8}: \sum\nolimits _{u}R_{i,\left(u,v\right)}=1, \sum\nolimits _{u}Q_{i,\left(u,v\right)}=N_{\mathrm{hop},i}, v=t_{i},
\end{align}
where constraint C5 ensures $s_{i}$ is the source node and constraint C6 ensures the flow originates from $s_{i}$; and constraint C7 indicates $t_{i}$ is the destination node and constraint C8 indicates the flow is destined for $t_{i}$. The routing matrix and the hop matrix also have the following constraint:
\begin{align}
\mathrm{C9}:Q_{i,\left(u,v\right)}=R_{i,\left(u,v\right)} \cdot Q_{i,\left(u,v\right)},\forall\left(u,v\right)\in\mathcal{E}.
\end{align}

In addition, the continuity of the flow should be satisfied:
\begin{align}
\mathrm{C10}:\sum\nolimits _{v}Q_{i,\left(u,v\right)}=\sum\nolimits _{v}Q_{i,\left(v,u\right)}+\sum\nolimits _{v}R_{i,\left(u,v\right)},
\end{align}
where $\ensuremath{u\in\mathcal{V}-\left\{ t_{i}\right\} ,\left(u,v\right)\in\mathcal{E}}$ and $\left(v,u\right)\in\mathcal{E}$. C10 ensures that the flow arriving at node $u$ at the $m$-th hop should leave at the $(m\!+\!1)$-th hop.
\subsubsection{VNF Placement Constraint}
Since the SFC considered is one-way and single-flow, any VNF in the SFC $\mathbb{F}_{i}$ of each service request $S_{i}$ can be only placed at one node, thus
\begin{align}
\mathrm{C11}:\sum\nolimits _{n}P_{i,(j,n)}=1,n\in\mathcal{V},
\end{align}
and considering all VNFs need to be placed, we have
\begin{align}
\mathrm{C12}:\sum\nolimits _{j}\sum\nolimits _{n}P_{i,(j,n)}=N_{\mathrm{f},i},n\in\mathcal{V},\ensuremath{j\in\left\{ 1,\ldots,N_{\mathrm{f},i}\right\} }.
\end{align}

To ensure the placed VNFs are part of the routing chain of $S_{i}$, we have
\begin{align}
&\mathrm{C13}:P_{i,(j,n)} \leq \sum\nolimits _{u}R_{i,\left(u,n\right)},n\in\mathcal{V}-\left\{ s_{i}\right\} ,u\in\mathcal{V}.\\
&\mathrm{C14}:P_{i,(j,n)}\leq\sum\nolimits _{v}R_{i,\left(n,v\right)},n\in\mathcal{V}-\left\{ t_{i}\right\} ,v\in\mathcal{V}.
\end{align}
If the $j$-th VNF requested by service request $S_{i}$ is placed at node $n$ but node $n$ is not part of the routing chain of $S_{i}$, then $P_{i,(j,n)}=1$ and $\sum\nolimits _{u}R_{i,\left(u,n\right)}=0$, i.e., $P_{i,(j,n)}>\sum\nolimits _{u}R_{i,\left(u,n\right)}$, does not satisfy constraint C13. For the remaining three cases, $P_{i,(j,n)}\leq\sum\nolimits _{u}R_{i,\left(u,n\right)}$ is valid and satisfies constraint C13, as follows: the $j$-th VNF is placed at node $n$, and node $n$ is part of the routing chain of $S_{i}$ (i.e., $P_{i,(j,n)}=1$ and $\sum\nolimits _{u}R_{i,\left(u,n\right)}=1$); the $j$-th VNF is not placed at node $n$, but node $n$ is part of the routing chain of $S_{i}$ (i.e., $P_{i,(j,n)}=0$ and $\sum\nolimits _{u}R_{i,\left(u,n\right)}=1$); and the $j$-th VNF is not placed at node $n$, and node $n$ is not part of the routing chain of $S_{i}$ (i.e., $P_{i,(j,n)}=0$ and $\sum\nolimits _{u}R_{i,\left(u,n\right)}=0$). Similar to C13, we can prove C14.

The required VNFs also need to be traversed in the correct order of the routing chain. In other words, the hop number corresponding to the $(j+1)$-th VNF in $\mathbb{F}_{i}$ is not smaller than the hop number corresponding to the $j$-th VNF:
\begin{align}
\mathrm{C15}:\sum\nolimits _{u}Q_{i,(u,v_{j}^{\ast})}\leq\sum\nolimits _{u}Q_{i,(u,v_{j+1}^{\ast})},
\end{align}
where $v_{j}^{\ast}$ is the node placed the $j$-th VNF of $S_{i}$, and $j\in\{1,\ldots,N_{\mathrm{f},i}-1\}$.
\subsubsection{Resource Constraints}
For each node $n$, the computing capacity and storage space requirements of all VNFs must not exceed the resource upper limit of the node itself, i.e.,
\begin{align}
&\mathrm{C16}:\sum\nolimits _{i}\sum\nolimits _{j}f_{\varGamma(P_{i,(j,n)})}^{\mathrm{c}}P_{i,(j,n)}\leq C_{n}, \\ &\mathrm{C17}:\sum\nolimits _{i}\sum\nolimits _{j}f_{\varGamma(P_{i,(j,n)})}^{\mathrm{m}}P_{i,(j,n)}\leq M_{n},
\end{align}
where $\varGamma(P_{i,(j,n)})$ stands for the category of the $j$-th VNF placed at node $n$ in the service $S_{i}$, $\varGamma(P_{i,(j,n)})\in\{1,\ldots,N_{\mathrm{f}}^{\mathrm{total}}\}$; and $f_{\varGamma(P_{i,(j,n)})}^{\mathrm{c}}$ can be further expressed as $f_{\varGamma(P_{i,(j,n)})}^{\mathrm{c}}=\eta_{\varGamma(P_{i,(j,n)})}r_{i}^{\mathrm{s}}$.

For each link $(u,v)$, the link bandwidth constraint can be expressed by
\begin{align}
\mathrm{C18}:\sum\nolimits _{i}r_{i}^{\mathrm{s}}R_{i,\left(u,v\right)}\leq B_{\left(u,v\right)}.
\end{align}
\subsection{Problem Formulation for Joint Minimizing Cost and Delay}
The cost and delay are minimized jointly in this paper. As per the cost, we consider both the deployment cost and resource utilization cost. Let $c_{k,n}^{\mathrm{d}}$ denote the deployment cost of the $k$-th category VNF at the node $n$. To calculate the resource utilization cost, we let $c_{n}^{\mathrm{c}}$ and $c_{n}^{\mathrm{m}}$ denote the cost per unit of computing and memory resources consumed at node $n$. $c_{(u,v)}^{\mathrm{B}}$ denotes the cost per unit of bandwidth resources consumed at link $(u,v)$. As a result, we have
\begin{align}
\resizebox{.85\hsize}{!}{$C_{\mathrm{total},i}=\underset{\forall n}{\sum}\underset{\forall j}{\sum}c_{\varGamma(P_{i,(j,n)}),n}^{\mathrm{d}}P_{i,(j,n)}+\underset{\forall n}{\sum}\underset{\forall j}{\sum}P_{i,(j,n)}\left(f_{\varGamma(P_{i,(j,n)})}^{\mathrm{c}}c_{n}^{\mathrm{c}}+f_{\varGamma(P_{i,(j,n)})}^{\mathrm{m}}c_{n}^{\mathrm{m}}\right)+\underset{\forall\left(u,v\right)}{\sum}r_{i}^{\mathrm{s}}R_{i,\left(u,v\right)}c_{(u,v)}^{\mathrm{B}}.$}
\end{align}

The service delay includes the transmission and processing delays. The processing delay is composed of the invariant processing delay of the node and the dynamic processing delay of different VNFs (depending on $r^{\mathrm{s}}_{i}$). Therefore, the delay is given by
\begin{align}
\resizebox{.85\hsize}{!}{$D_{\mathrm{total},i}=\underset{\forall\left(u,v\right)}{\sum}r_{i}^{\mathrm{s}}R_{i,\left(u,v\right)}d_{(u,v)}^{\mathrm{B}}+d^{\mathrm{inv}}\left(1+\underset{\forall u}{\sum}\underset{\forall v}{\sum}R_{i,\left(u,v\right)}\right)+\underset{\forall n}{\sum}\underset{\forall j}{\sum}P_{i,(j,n)}\left(d_{\varGamma(P_{i,(j,n)})}^{\mathrm{dyn}}r_{i}^{\mathrm{s}}\right),$}
\end{align}
where $d_{(u,v)}^{\mathrm{B}}$ is the delay of link $(u,v)$ per unit service rate, $d^{\mathrm{inv}}$ is the invariant processing delay of each node, and $d_{\varGamma(P_{i,(j,n)})}^{\mathrm{dyn}}$ is the dynamic processing delay of the $\varGamma(P_{i,(j,n)})$-th category VNF. $\ensuremath{1+\sum_{u}\sum_{v}R_{i,\left(u,v\right)}}$ is the number of nodes along the routing chain of $S_{i}$.

The cost and delay, as two key feature requirements in service requests \cite{8464883,8626768}, also have general differences among different users. For supporting the differentiated service experience, some users are considered delay-sensitive, while some users are considered cost-sensitive\cite{8459956}. Therefore, we propose a demand-aware factor $(\varphi_{i}^{\mathrm{c}},\varphi_{i}^{\mathrm{d}})$ ($\varphi_{i}^{\mathrm{c}}+\varphi_{i}^{\mathrm{d}}=1$, $\varphi_{i}^{\mathrm{c}},\varphi_{i}^{\mathrm{d}}\!\in\!\left[0,1\right]$) to weight the cost and delay for user $i$. The larger $\varphi_{i}^{\mathrm{c}}$ is, the more sensitive user $i$ is to the cost. The larger $\varphi_{i}^{\mathrm{d}}$ is, the more sensitive user $i$ is to the delay\footnote{$\varphi_{i}^{\mathrm{c}}$ and $\varphi_{i}^{\mathrm{d}}$ can be obtained in two methods: 1) Passive method. By analyzing the historical behavior data of each user, the demand preference can be easily obtained; and 2) Active method. The users can directly report personal demand preferences.}. We minimize the weighted sum of cost and delay of all service requests at the same time. The optimization problem\footnote{Due to resource constraints, when multiple requests are optimized at the same time, the feasible region simultaneously meeting constraints C1--C18 may not exist, so that no service request can be accepted. To this end, this paper has two key designs: 1) Batch-by-batch service request deployment mechanism (service requests within each batch are deployed in parallel); and 2) Limit the number of episodes about exploring the feasible region. Please refer to Section V-C for more details.} is formulated as:
\begin{align}
OP1:\quad\underset{P_{i,(j,n)},R_{i,\left(u,v\right)}}{\min}&\stackrel[i=1]{M}{\sum}\left(\varphi_{i}^{\mathrm{c}}\psi^{\mathrm{c}}C_{\mathrm{total},i}+\varphi_{i}^{\mathrm{d}}\psi^{\mathrm{d}}D_{\mathrm{total},i}\right) \nonumber \\
&\mbox{s.t.} \quad \mathrm{C}1-\mathrm{C}18,
\end{align}
where $C_{\mathrm{total},i}$ and $D_{\mathrm{total},i}$ are dimensionless and counted by ``Unit''. $\psi^{\mathrm{c}}$ and $\psi^{\mathrm{d}}$ are used to eliminate the order of magnitude difference in the values of $C_{\mathrm{total},i}$ and $D_{\mathrm{total},i}$ caused by the dimensional influence. The VNF placement problem is a sub-problem of \emph{OP1}, and the NP-completeness of VNF placement has been proved in \cite{lin2016demand}, our \emph{OP1} is also NP-complete. \emph{OP1} is mathematically intractable and computationally prohibitive.
\section{Proposed Reinforcement Learning-Based Approach}
\begin{figure*}[hbt]
\centering{}\includegraphics[scale=0.7]{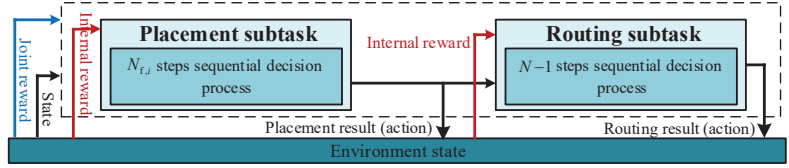}
\caption{Transformation of optimization problem \emph{OP1}.}\label{resource_stru_sample_new}
\end{figure*}

Problem \emph{OP1} involves the joint optimization of $P_{i,(j,n)}$ and $R_{i,(u,v)}$, which interplay with each other. To simplify the neural network structure and reduce the action space to benefit convergence, as shown in Fig. \ref{resource_stru_sample_new}, \emph{OP1} is decomposed into two iterative subtasks, including a routing subtask responsible for the \emph{VNF routing} and a placement subtask responsible for the \emph{VNF placement}. When solving \emph{OP1}, we first perform the placement subtask and obtain the placement result. Then, the routing subtask is performed, according to the placement result. Since the placement and routing subtasks affect each other, after some iterations, the optimal solution can be obtained. For efficiently solving placement and routing subtasks, RL is employed. A standard RL process can be defined as a Markov process. At each decision step $t$, the agent chooses an \textbf{action} $a_{t}\in\mathcal{A}$ from the target policy $\pi$ based on the \textbf{state} $s_{t}\in\mathcal{S}$, and executes it in the environment $E$, then a \textbf{reward} $r_{t}$ can be received. The agent interacts with the environment $E$ to maximize the expected future rewards, which are discounted by a factor $\gamma$ per step. The discounted reward is $\mathcal{R}_{t}=\sum_{t^{'}=t}^{T}\gamma^{(t^{'}-t)}r_{t^{'}}$ for step $t$, where $T$ is the number of steps at which RL task is terminated.

This paper studies the simultaneous deployment of $M$ service requests, and thus the placement and routing subtasks are transformed into $M$ concurrent parallel sequential decision processes. $2M$ agents are needed, including $M$ placement agents and $M$ routing agents. The advantages of using multi-agent technique include: 1) Facilitate the transformation of \emph{OP1} into sequential decision processes. The single-agent RL is difficult to transform simultaneously $M$ service requests into one sequence decision process, considering the differentiated VNF demands of different services; and 2) Facilitate the design of placement and routing actions. When adopting the single-agent technique, the one-hot encoding-based action requires a total of $\prod_{i}(\frac{N!}{(N-N_{\mathrm{f},i})!})$ output neurons. It is also impossible to output actions for each user at the same time by grouping the output neurons, due to the different number of VNFs required by different service requests.

In the placement subtask, the sequential decision process of each request consists of $N_{\mathrm{f}}^{\mathrm{max}}$ decision steps ($N_{\mathrm{f}}^{\mathrm{max}}$ is the maximum number of VNFs required by all services): At each step $t$, the placement agent outputs the placement node of a VNF in $\mathbb{F}_{i}$ of $S_{i}$. The process terminates after all required VNFs are placed (the process has an early termination when $N_{\mathrm{f},i}\!<\!\ensuremath{N_{\mathrm{f}}}^{\mathrm{max}}$). In the routing subtask, the sequential decision process of each $S_{i}$ consists of $N-1$ steps. The starting routing node of the sequential decision process is the source node $s_{i}$, and each subsequent decision step outputs a routing node. This process terminates after $N\!-\!1$ steps. If the output of an intermediate decision step is the destination node $r_{i}$, the process has an early termination.
\subsection{Action}
Let $\mathbf{a}_{i,t}^{\mathrm{p}}=[\rho_{i,1,t},\ldots,\rho_{i,n,t},\ldots,\rho_{i,N,t}]_{1\times N}$ ($\rho_{i,n,t}\in\left\{ 0,1\right\}$) be the action of the $i$-th placement agent for the decision step $t$. If node $n$ is selected to place the VNF at decision step $t$, $\rho_{i,n,t}=1$; otherwise, $\rho_{i,n,t}=0$. The final placement result of $S_{i}$ denoted by $\mathbf{P}_{i}$, is given by
\begin{align}\label{self_action_to_result}
\mathbf{P}_{i}=[(\mathbf{a}_{i,1}^{\mathrm{p}})^{\mathrm{T}},\ldots,(\mathbf{a}_{i,t}^{\mathrm{p}})^{\mathrm{T}},\ldots,(\mathbf{a}_{i,N_{\mathrm{f},i}}^{\mathrm{p}})^{\mathrm{T}}]_{N_{\mathrm{f},i}\times N}^{\mathrm{T}}.
\end{align}

Let $\mathbf{a}_{i,t}^{\mathrm{r}}=[\xi_{i,1,t},\ldots,\xi_{i,n,t},\ldots,\xi_{i,N,t}]_{1\times N}$ ($\xi_{i,n,t}\in\left\{ 0,1\right\}$) be the action of the $i$-th routing agent at step $t$. $\xi_{i,n,t}=1$ if node $n$ is selected as a routing node of $S_{i}$ at step $t$; $\xi_{i,n,t}=0$, otherwise.
\subsection{State}
In the proposed multi-agent RL, the state of each agent includes observations made by the other agents and additional state information (if available) to increase the learning efficiency \cite{lowe2017multi}. The state of the $i$-th placement agent for decision step $t$, i.e., $s_{i,t}^{\mathrm{p}}$, is given by
\begin{align}
\resizebox{.85\hsize}{!}{$\left\{ \underset{\mathrm{self-observation}}{\underbrace{r_{i}^{\mathrm{s}},\varphi_{i}^{\mathrm{c}},\varphi_{i}^{\mathrm{d}},s_{i},t_{i},\mathbb{F}_{i},\mathbb{F}_{i,t}^{\mathrm{re}}}},\underset{\mathrm{other-observation}}{\underbrace{r_{i}^{\mathrm{s}},\varphi_{i}^{\mathrm{c}},\varphi_{i}^{\mathrm{d}},s_{i},t_{i},\mathbb{F}_{1},\ldots,r_{M}^{\mathrm{s}},\varphi_{M}^{\mathrm{c}},\varphi_{M}^{\mathrm{d}},s_{M},t_{M},\mathbb{F}_{M}}},\underset{\mathrm{SN-information}}{\underbrace{\mathbf{C}_{t}^{\mathrm{re}},\mathbf{M}_{t}^{\mathrm{re}},\mathbf{A}}}\right\} $,}
\end{align}
where $s_{i,t}^{\mathrm{p}}$ consists of three parts: the $i$-th agent's own observation, the other agents' observations, and the additional information of the SN (i.e., the resource state and network topology). In particular, $\mathbb{F}_{i,t}^{\mathrm{re}}=[\mathscr{F}_{i,1,t}^{\mathrm{re}},\ldots,\mathscr{F}_{i,j,t}^{\mathrm{re}},\ldots,\mathscr{F}_{i,N_{\mathrm{f},i},t}^{\mathrm{re}}]_{1\times N_{\mathrm{f},i}}$, $\mathscr{F}_{i,j,t}^{\mathrm{re}}\in\left\{ 0,1\right\}$. If the $j$-th VNF of $S_{i}$ completes the placement at step $t$, $\mathscr{F}_{i,j,t}^{\mathrm{re}}=0$; otherwise, $\mathscr{F}_{i,j,t}^{\mathrm{re}}=1$. $\mathbf{C}_{t}^{\mathrm{re}}=\left[C_{1,t}^{\mathrm{re}},\ldots,C_{n,t}^{\mathrm{re}},\ldots,C_{N,t}^{\mathrm{re}}\right]_{1\times N}$ with $C_{n,t}^{\mathrm{re}}$ being the remaining computing capacity available for node $n$ at step $t$. $\mathbf{M}_{t}^{\mathrm{re}}=[M_{1,t}^{\mathrm{re}},\ldots,M_{n,t}^{\mathrm{re}},
\ldots,M_{N,t}^{\mathrm{re}}]_{1\times N}$ with $M_{n,t}^{\mathrm{re}}$ being the remaining memory size available for node $n$ at step $t$ Likewise, the routing state of the $i$-th routing agent for step $t$, $s_{i,t}^{\mathrm{r}}$, is given by
\begin{align}\label{routing_state}
\resizebox{.85\hsize}{!}{$\ensuremath{\left\{ \underset{\mathrm{self-observation}}{\underbrace{r_{i}^{\mathrm{s}},\varphi_{i}^{\mathrm{c}},\varphi_{i}^{\mathrm{d}},s_{i},t_{i},\mathbf{P}_{i},\mathbf{R}_{i,t}^{\mathrm{cu}}}},\underset{\mathrm{other-observation}}{\underbrace{r_{1}^{\mathrm{s}},\varphi_{1}^{\mathrm{c}},\varphi_{1}^{\mathrm{d}},s_{1},t_{1},\mathbf{P}_{1},\ldots,r_{M}^{\mathrm{s}},\varphi_{M}^{\mathrm{c}},\varphi_{M}^{\mathrm{d}},s_{M},t_{M},\mathbf{P}_{M}}},\underset{\mathrm{SN-information}}{\underbrace{\mathbf{B}_{t}^{\mathrm{re}},\mathbf{A}}}\right\} },$}
\end{align}
where $\mathbf{B}_{t}^{\mathrm{re}}=[B_{(u,v),t}^{\mathrm{re}}]_{N\times N}$ with $B_{(u,v),t}^{\mathrm{re}}$ the remaining bandwidth of link $(u,v)$ at step $t$.

The \emph{current routing matrix} $\mathbf{R}_{i,t}^{\mathrm{cu}}=[R_{i,(u,v),t}^{\mathrm{cu}}]_{N\times N}$ captures the routing path output by the previous $(t-1)$ steps of the $i$-th routing agent. If link $(u,v)$ is part of
 the routing path, $R_{i,(u,v),t}^{\mathrm{cu}}=1$; otherwise, $R_{i,(u,v),t}^{\mathrm{cu}}=0$. $\mathbf{R}_{i,t}^{\mathrm{cu}}$ changes with the routing action $\mathbf{a}_{i,t}^{\mathrm{r}}$ at each step $t$, i.e.,
\begin{align}\label{routing_action_result}
\mathbf{R}_{i,t}^{\mathrm{cu}}=\mathbf{R}_{i,t-1}^{\mathrm{cu}}+(\mathbf{a}_{i,t-2}^{\mathrm{r}})^{\mathrm{T}}\mathbf{a}_{i,t-1}^{\mathrm{r}},
\end{align}
where $t\geq2$. For $t=1$, $\mathbf{R}_{i,1}^{\mathrm{cu}}=\mathbf{0}_{N\times N}$ and $\mathbf{a}_{i,0}^{\mathrm{r}}=[\pi_{i,1,t},\ldots,\pi_{i,n,t},\ldots,\pi_{i,N,t}]_{1\times N}$. If node $n$ is the source node $s_{i}$, $\pi_{i,n,t}=1$; otherwise, $\pi_{i,n,t}=0$.
\subsection{Reward Function Design}
As compared with a traditional RL task, the multi-agent RL task studied in this paper is challenging: 1) The placement and routing agents have to learn in an alternating manner; and 2) Abundant constraints need to be satisfied. Inspired by the typical work in \cite{bello2016neural} and the application of ``pruning'' algorithm in RL \cite{livne2020pops}, we design a novel \emph{joint reward} and \emph{internal reward} mechanism, where the joint reward is for constraint satisfaction and the joint reward is for achieving optimization objectives. Compared with unifying constraints and objectives into one reward expression, our internal reward terminates the action exploration in the non-feasible region early, which is equivalent to ``pruning'' the invalid exploration process.
\begin{table*}[htb]
	\centering{}
	\textbf{Table \uppercase\expandafter{\romannumeral2}}~~The summary of constraints that have been met.\\
\small
\setlength{\tabcolsep}{1.850mm}{
		\begin{tabular}{|c|l|c|l|}
        \hline
        \multirow{4}*{\tabincell{c}{C1\\C2\\C4\\C9}} & \multirow{4}*{\tabincell{c}{Each routing agent\\ selects only one routing\\ node at any decision\\ step $t$, i.e., $\sum_{n}\xi_{i,n,t}=1$}} &  \multirow{4}*{\tabincell{c}{C5\\C6\\C10\\C12}}& \multirow{2}*{\tabincell{c}{Any node $s_{i}$ is preset to the starting node of each service flow $S_{i}$}}\\
        ~ & ~&~&~ \\
        \cline{3-4}
        ~ & ~ &~ & The routing subtask is a sequential decision process \\
        \cline{3-4}
        ~ & ~ &~ & Each placement subtask is an $N_{\mathrm{f},i}$-step sequential decision process\\
        \hline
        \multirow{2}*{\tabincell{c}{C7\\C8}} & \multicolumn{3}{c|}{\multirow{2}*{\tabincell{c}{During the learning process of each routing gent, the destination node $r_{i}$ is checked all the time. \\Once the destination node $r_{i}$ is selected, the learning process is early terminated.}}}\\
      ~&\multicolumn{3}{c|}{\multirow{2}*{\tabincell{c}{~}}}\\
    \hline
        C11 & \multicolumn{3}{c|}{\tabincell{c}{Each placement agent selects only one node at any decision step $t$, i.e., $\sum_{n}\rho_{i,n,t}=1$}}\\
        \hline
	\end{tabular}}
\end{table*}

Since C1, C2, and C4--C12 have been met during the transformation of subtasks and the definition of the action space, as shown in Table \uppercase\expandafter{\romannumeral2}, the placement agents only need to satisfy C16 and C17. The routing agents only need to satisfy C3, C13--C15, and C18. The placement results that satisfy C16 and C17 are referred to as suitable placement results (SPRs), and those that do not satisfy the constraints are referred to as non-SPRs. Likewise, suitable routing results (SRRs) and non-SRRs indicate whether the routing results satisfy C3, C13--C15, and C18 or not. For the $i$-th placement agent, the internal reward $r_{i,t}^{\mathrm{p,int}}$ is defined as
\begin{align}\label{inter_placement}
r_{i,t}^{\mathrm{p,int}}=\omega_{i}^{\mathrm{p,int}}\left(\varUpsilon\left(\mathrm{C}16\right)+\varUpsilon\left(\mathrm{C}17\right)\right),
\end{align}
where $\omega_{i}^{\mathrm{p,int}}<0$ is a penalty factor. $\varUpsilon\left(\mathrm{C}x\right)$ is a discriminant function. If constraint C$x$ is met, $\varUpsilon\left(\mathrm{C}x\right)\!=\!0$; otherwise, $\varUpsilon\left(\mathrm{C}x\right)\!=\!1$. In the routing agents, we divide the constraints into three classes, according to the priority of different constraints: Class I (C18); Class II (C13--C15); and Class III (C3). The internal reward $r_{i,t}^{\mathrm{r,int}}$ of the $i$-th routing agent is
\begin{align}\label{inter_routing}
\resizebox{.85\hsize}{!}{$r_{i,t}^{\mathrm{r,int}}=\omega_{\mathrm{I},i}^{\mathrm{r,int}}\varUpsilon\left(\mathrm{C}18\right)+\omega_{\mathrm{II},i}^{\mathrm{r,int}}\sum_{x}\varUpsilon\left(\mathrm{C}x\right)+\omega_{\mathrm{III},i}^{\mathrm{r,int}}\varUpsilon\left(\mathrm{C}3\right),
$}
\end{align}
where $x\in \{13,14,15\}$; and $\omega_{\mathrm{I},i}^{\mathrm{r,int}}$, $\omega_{\mathrm{II},i}^{\mathrm{r,int}}$ and $\omega_{\mathrm{III},i}^{\mathrm{r,int}}$ are weighting coefficients.

The joint rewards of the $i$-th placement and routing agents, $\ensuremath{r_{i,t}^{\mathrm{p,jo}}}$ and $\ensuremath{r_{i,t}^{\mathrm{r,jo}}}$, depend on the objective of \emph{OP1}. However, the joint rewards cannot directly use the objective value of \emph{OP1}: 1) The objective value of \emph{OP1} is the same for each agent, making it difficult for each agent to infer its individual contribution to the global objective \cite{foerster2018counterfactual}; and 2) the demand-aware factors of different users (or agents) may be different, but the global reward does not motivate the users to maintain the individuals' difference. By employed a ``difference reward'' technique \cite{tumer2007distributed}, we decouple the total optimization objective and define the joint reward as
\begin{align}\label{joint_reward}
\ensuremath{r_{i,t}^{\mathrm{p,jo}}}=\ensuremath{r_{i,t}^{\mathrm{r,jo}}}=&\eta_{i}^{\mathrm{dec}}\omega_{\mathrm{dec}}^{\mathrm{jo}}\times\exp(\omega_{\mathrm{scal}}^{\mathrm{jo,\mathrm{exp}}}(\sum_{i=1}^{M}(\varphi_{i}^{\mathrm{c}}\psi^{\mathrm{c}}C_{\mathrm{total},i}\nonumber\\
&+\varphi_{i}^{\mathrm{d}}\psi^{\mathrm{d}}D_{\mathrm{total},i}))+\omega_{\mathrm{trans}}^{\mathrm{jo,\mathrm{exp}}}),
\end{align}
where $\eta_{i}^{\mathrm{dec}}=(\Pi_{z=1(z\neq i)}^{M}\Theta_{z})/\sum_{i}^{M}(\Pi_{z=1(z\neq i)}^{M}\Theta_{z})$; $\Theta_{z}=\varphi_{z}^{\mathrm{c}}\psi^{\mathrm{c}}C_{\mathrm{total},z}+\varphi_{z}^{\mathrm{d}}\psi^{\mathrm{d}}D_{\mathrm{total},z}$; and $\omega_{\mathrm{dec}}^{\mathrm{jo}}>0$, $\omega_{\mathrm{scal}}^{\mathrm{jo,\mathrm{exp}}}\!<\!0$, and $\omega_{\mathrm{trans}}^{\mathrm{jo,\mathrm{exp}}}\!>\!0$ are adjustable coefficients. The joint reward uses a negative-exponential function inspired by the ``reward shaping'' technique \cite{ng1999policy}. The negative exponential changes of the reward incentivize the agents to increasingly fast approach their optimization objective. The reward grows exponentially, and hence increasingly faster, as the objective decreases.
\subsection{Theoretical Preliminaries of DDPG}
RL algorithms widely uses the action-value function $Q(s_{t},a_{t})$ (here, $s_{t}$ and $a_{t}$ are $s_{i,t}^{\mathrm{p}}$ and $\mathbf{a}_{i,t}^{\mathrm{p}}$ in the case of placement, or $s_{i,t}^{\mathrm{r}}$ and $\mathbf{a}_{i,t}^{\mathrm{r}}$ in the case of routing) to describe the expected reward for choosing the action $a_{t}$ in the state $s_{t}$. Let $\pi$ denote the target policy distribution (stochastic or deterministic), $Q(s_{t},a_{t})$ can be calculated by $\mathbb{E}_{r_{t},s_{t}\sim E,a_{t}\sim\pi}[\mathcal{R}_{t}|s_{t},a_{t}]$. Further, let $\mu$ be the deterministic policy, $\mu$: $\mathcal{S}\leftarrow\mathcal{A}$, then $Q(s_{t},a_{t})$ with the bellman equation can be written as
\begin{align}\label{Q_claculate_simple}
Q^{\mu}\left(s_{t},a_{t}\right)=\mathbb{E}_{r_{t},s_{t+1}\sim E}[r_{t}+\gamma Q^{\mu}\left(s_{t+1},\mu\left(s_{t+1}\right)\right)].
\end{align}

In DRL, the $Q\left(\cdot\right)$ function can be approximated by using a deep neural network (DNN). A DNN is also known to cause instability or even divergence. To address this issue, DQN, a typical DRL algorithm, uses the \emph{target network} and \emph{experience relay} techniques \cite{mnih2015human}. Nevertheless, DQN is still value-based learning, and is usually is suitable for discrete and low-dimensional actions \cite{lillicrap2015continuous}. DDPG, drawing on the respective advantages the actor-critic (AC) structure and DQN, can effectively address the challenge resulting from high-dimensional or continuous action spaces \cite{silver2014deterministic}. Compared with other DRL algorithms, DDPG has the following advantages: 1) Multi-agent DDPG algorithm has been widely verified as a stable and reliable multi-agent algorithm \cite{lowe2017multi}; 2) More suitable for our research problem characteristic\footnote{Since the problem considered has numerous constraints, in each decision, a deterministic deployment policy that can realize the user's personalized service demands under a given resource is required.}; 3) More simpler structure compared to other deterministic policy algorithms; and 4) More suitable for large-scale network topology decisions compared with other value-based algorithms such as dueling DQN (DDQN).

The AC structure includes a critic network (i.e., Q-network) and an actor network (i.e., policy network). Each critic network and actor network has a copy network, where the copy network is called the target network and the original network is called the online network. Let $\theta^{Q}$ and $\theta^{Q^{'}}$ be the parameters of the online and target critic networks respectively, and let $\theta^{\mu}$ and $\theta^{\mu^{'}}$ be the parameters of the online and target actor networks respectively. The online critic and actor networks are trained online, and their parameters ($\theta^{Q}$ and $\theta^{\mu}$) are updated by gradient. The target critic and actor networks are updated by utilizing the ``soft update'' algorithm \cite{silver2014deterministic}.
\section{Proposed MADRL-P\&R Framework}
\subsection{Implementation of MADRL-P\&R Framework}
\begin{figure}[htb]
\centering{}\includegraphics[scale=0.34]{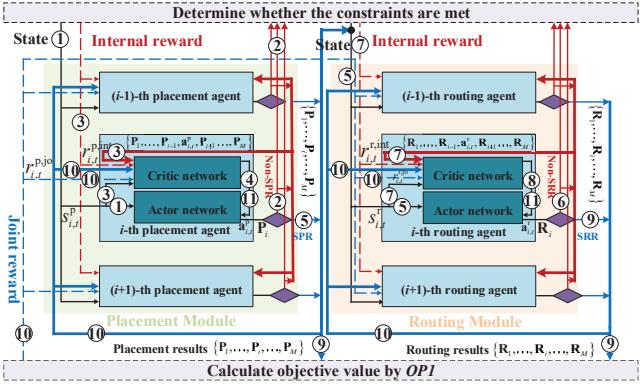}
\caption{MADRL-P\&R framework. ``Thick red lines'' are non-SPRs or non-SRRs, and ``thick blue lines'' are SPRs or SRRs,}\label{globe_framework}
\end{figure}

Based on the DDPG model, the proposed global MADRL-P\&R framework is implemented in Fig. \ref{globe_framework}. The framework\footnote{Our multi-agent framework is centrally deployed on an ``online resource allocator'' at the application layer, as shown in the system architecture in \cite{8480442}. Due to the centralized deployment, different agents can achieve agent synchronization through the internal control of the algorithm without additional communication overhead.} consists of a multi-agent placement module and a multi-agent routing module. For achieving the optimal policy and ensuring the stable convergence, the critic network of each agent needs to consider the actions (or policies) of all other agents \cite{kraemer2016multi}.

\textbf{Operation mechanism of MADRL-P\&R framework:} A complete placement and routing iteration round is referred to an ``epoch''. In each epoch, the operation process of the MADRL-P\&R framework includes steps \textcircled{\small{1}}--\textcircled{\small{11}}. In step \textcircled{\small{1}}, all placement agents execute the placement actor network (PAN) to output the placement action $\mathbf{a}_{i,t}^{\mathrm{p}}$, based on their individual observation states $s_{i,t}^{\mathrm{p}}$. After each placement agent executes $N_{\mathrm{f},i}$ steps, the placement result $\mathbf{P}_{i}$ is obtained by (\ref{self_action_to_result}). If all $\mathbf{P}_{i}$ are SPRs, then steps \textcircled{\small{5}}--\textcircled{\small{11}} are performed; otherwise, steps  \textcircled{\small{2}}--\textcircled{\small{4}} are performed.

Steps \textcircled{\small{2}}--\textcircled{\small{4}}. For non-SPRs, step \textcircled{\small{2}} assesses whether the placement action $\mathbf{a}_{i,t}^{\mathrm{p}}$ of each step satisfies constraints C16 and C17, and then calculates the internal reward $r_{i,t}^{\mathrm{p,int}}$ of each step by (\ref{inter_placement}). In step \textcircled{\small{3}}, $r_{i,t}^{\mathrm{p,int}}$, $s_{i,t}^{\mathrm{p}}$, $\mathbf{a}_{i,t}^{\mathrm{p}}$ and the non-SPRs $\ensuremath{\{\mathbf{P}_{1},\ldots,\mathbf{P}_{i-1},\mathbf{P}_{i+1}\ldots,\mathbf{P}_{M}\}}$ of the other agents are entered into the placement critic network (PCN) for calculating corresponding action-value function by (\ref{Q_claculate_simple}). In step \textcircled{\small{4}}, according to the action-value function, the PAN is updated by the policy gradient method. Steps \textcircled{\small{1}}--\textcircled{\small{4}} repeat until the placement agents generate SPRs.

In step \textcircled{\small{5}}, all SPRs $\{ \mathbf{P}_{1},\mathbf{P}_{2},\ldots,\mathbf{P}_{M}\} $ are part of the placement state $s_{i,t}^{\mathrm{r}}$. Based on $s_{i,t}^{\mathrm{r}}$, all routing agents execute the routing actor network (RAN) to output the routing action $\mathbf{a}_{i,t}^{\mathrm{r}}$. After each agent executes at most $(N\!-\!1)$ steps, the routing result $\mathbf{R}_{i}$ is obtained by (\ref{routing_action_result}). If all $\mathbf{R}_{i}$ are SRRs, then steps \textcircled{\small{9}}--\textcircled{\small{11}} are performed; otherwise, steps \textcircled{\small{6}}--\textcircled{\small{8}} are performed.

Steps \textcircled{\small{6}}--\textcircled{\small{8}}. For non-SRR, step \textcircled{\small{6}} assesses whether the routing action $\mathbf{a}_{i,t}^{\mathrm{r}}$ of each step satisfies constraints C3, C13--C15 and C18, and calculates the internal reward $r_{i,t}^{\mathrm{r,int}}$ of each step by (\ref{inter_routing}). In step \textcircled{\small{7}}, $r_{i,t}^{\mathrm{r,int}}$, $s_{i,t}^{\mathrm{r}}$, $\mathbf{a}_{i,t}^{\mathrm{r}}$ and the non-SRRs $\ensuremath{\{\mathbf{R}_{1},\ldots,\mathbf{R}_{i-1},\mathbf{R}_{i+1}\ldots,\mathbf{R}_{M}\}}$ of the other agents are input into the routing critic network (RCN) to calculate the action-value function. In step \textcircled{\small{8}}, according to the action-value function, the RAN is updated by the policy gradient method. Steps \textcircled{\small{6}}--\textcircled{\small{8}} repeat until the placement agents generate SRRs.

Steps \textcircled{\small{9}}--\textcircled{\small{11}}. In step \textcircled{\small{9}}, we substitute ($\{ \mathbf{P}_{1},\mathbf{P}_{2},\ldots,\mathbf{P}_{M}\}$ and $\{ \mathbf{R}_{1},\mathbf{R}_{2},\ldots,\mathbf{R}_{M}\}$) into (21) for calculating corresponding objective value, and then based on (\ref{joint_reward}), the joint reward $r_{i,t}^{\mathrm{p,jo}}$ and $r_{i,t}^{\mathrm{r,jo}}$ can be obtained. In step \textcircled{\small{10}}, $r_{i,t}^{\mathrm{p,jo}}$, $s_{i,t}^{\mathrm{p}}$, $\mathbf{a}_{i,t}^{\mathrm{p}}$ and the SPRs $\ensuremath{\{\mathbf{P}_{1},\ldots,\mathbf{P}_{i-1},\mathbf{P}_{i+1}\ldots,\mathbf{P}_{M}\}}$ of the other agents are input the PCN to calculate the action-value function. $r_{i,t}^{\mathrm{r,jo}}$,  $s_{i,t}^{\mathrm{r}}$, $\mathbf{a}_{i,t}^{\mathrm{r}}$ and the SRRs $\ensuremath{\{\mathbf{R}_{1},\ldots,\mathbf{R}_{i-1},\mathbf{R}_{i+1}\ldots,\mathbf{R}_{M}\}}$ are input to the RCN to calculate the action-value function. In step \textcircled{\small{11}}, according to the action-value function, the PAN anc RAN are further improved.
\subsection{Updating Algorithm and Network Structure}
In this paper, the \emph{centralized action-value function} \cite{lowe2017multi} is considered. We use the placement and routing results (i.e., $\mathbf{P}_{i}$ and $\mathbf{R}_{i}$) of the other agents, as opposed to the placement and routing actions (i.e., $\mathbf{a}_{i,t}^{\mathrm{p}}$ and $\mathbf{a}_{i,t}^{\mathrm{r}}$). This is because the decision steps required by each agent to obtain the placement or routing results are different. While some agents are making their decisions, the others already have obtained the placement or routing results in place. For the $i$-th placement agent, our centralized action-value function is
\begin{align}
Q_{i}^{\mathrm{p}}(s_{i,t}^{\mathrm{p}},\mathbf{a}_{i,t}^{\mathrm{p}},\mathbf{P}_{1},\ldots,\mathbf{P}_{i-1},\mathbf{P}_{i+1},\ldots,\mathbf{P}_{M}).
\end{align}
Based on the proposed centralized action-value function, for the PCN with parameters $\theta_{i}^{\mathrm{p},Q}$ of the $i$-th placement agent, the gradient can be calculated by
\begin{align}\label{multi_placement_critic}
&\nabla_{\theta_{i}^{\mathrm{p},Q}}L_{t}(\theta_{i}^{\mathrm{p},Q})=\mathbb{E}_{\mathbf{a}^{\mathrm{p},i}\sim\pi,s_{i}^{\mathrm{p}}\sim E}[(r_{i,t}^{\mathrm{p,fin}}\nonumber \\
&+\gamma Q_{i}^{\mathrm{p}}(s_{i,t+1}^{\mathrm{p}},\mu(s_{i,t+1}^{\mathrm{p}}),\mathbf{P}_{i}^{\mathrm{other}}|\theta_{i}^{\mathrm{p},Q^{'}})\\
&-Q_{i}^{\mathrm{p}}(s_{i,t}^{\mathrm{p}},\mu(s_{i,t}^{\mathrm{p}}),\mathbf{P}_{i}^{\mathrm{other}}|\theta_{i}^{\mathrm{p},Q}))\nabla_{\theta_{i}^{\mathrm{p},Q}}Q_{i}^{\mathrm{p}}(s_{i,t}^{\mathrm{p}},\mu(s_{i,t}^{\mathrm{p}}),\mathbf{P}_{i}^{\mathrm{other}}|\theta_{i}^{\mathrm{p},Q})].\nonumber
\end{align}
where $\mathbf{P}^{\mathrm{other}}\!=\!\{ \mathbf{P}_{1},\ldots,\mathbf{P}_{i-1},\mathbf{P}_{i+1},\ldots,\mathbf{P}_{M}\}$; $\theta_{i}^{\mathrm{p},Q}$ and $\theta_{i}^{\mathrm{p},Q^{'}}$ are the parameters of the online and target PCNs; the policy gradient of PAN with parameter $\theta_{i}^{\mathrm{p},\mu}$ of the $i$-th placement agent is
\begin{align}\label{multi_placement_actor}
&\nabla_{\theta_{i}^{\mathrm{p},\mu}}J_{i}^{\mathrm{p}}\nonumber\\
&\approx\mathbb{E}_{s_{i}^{\mathrm{p}}\sim E}\left[\nabla_{\theta_{i}^{\mathrm{p},\mu}}Q_{i}^{\mathrm{p}}(s_{i}^{\mathrm{p}},\mathbf{a}_{i}^{\mathrm{p}},\mathbf{P}_{i}^{\mathrm{other}}|\theta_{i}^{\mathrm{p},Q})|_{s_{i}^{\mathrm{p}}=s_{i,t}^{\mathrm{p}},\mathbf{a}_{i}^{\mathrm{p}}=\mu_{i}^{\mathrm{p}}(s_{i,t}^{\mathrm{p}}|\theta_{i}^{\mathrm{p},\mu})}\right]\nonumber\\
&=\mathbb{E}_{s_{i}^{\mathrm{p}}\sim E}[\nabla_{\mathbf{a}_{i}^{\mathrm{p}}}Q_{i}^{\mathrm{p}}(s_{i}^{\mathrm{p}},\mathbf{a}_{i}^{\mathrm{p}},\mathbf{P}_{i}^{\mathrm{other}}|\theta_{i}^{\mathrm{p},Q})|_{s_{i}^{\mathrm{p}}=s_{i,t}^{\mathrm{p}},\mathbf{a}_{i}^{\mathrm{p}}=\mu_{i}^{\mathrm{p}}(s_{i,t}^{\mathrm{p}})}\nonumber\\
&\nabla_{\theta_{i}^{\mathrm{p},\mu}}\mu_{i}^{\mathrm{p}}(s_{i}^{\mathrm{p}}|\theta_{i}^{\mathrm{p},\mu})|_{s_{i}^{\mathrm{p}}=s_{i,t}^{\mathrm{p}}}].
\end{align}

The update algorithms of the RCN and RAN are similar to (\ref{multi_placement_critic}) and (\ref{multi_placement_actor}), as given by
\begin{align}\label{multi_routing_critic}
&\nabla_{\theta_{i}^{\mathrm{r},Q}}L_{t}(\theta_{i}^{\mathrm{r},Q})=\mathbb{E}_{\mathbf{a}^{\mathrm{r},i}\sim\pi,s_{i}^{\mathrm{r}}\sim E}[(r_{i,t}^{\mathrm{r,fin}}\nonumber\\
&+\gamma Q_{i}^{\mathrm{r}}(s_{i,t+1}^{\mathrm{r}},\mu(s_{i,t+1}^{\mathrm{r}}),\mathbf{R}_{i}^{\mathrm{other}}|\theta_{i}^{\mathrm{r},Q^{'}})\\
&-Q_{i}^{\mathrm{r}}(s_{i,t}^{\mathrm{r}},\mu(s_{i,t}^{\mathrm{r}}),\mathbf{R}_{i}^{\mathrm{other}}|\theta_{i}^{\mathrm{r},Q}))\nabla_{\theta_{i}^{\mathrm{r},Q}}Q_{i}^{\mathrm{r}}(s_{i,t}^{\mathrm{r}},\mu(s_{i,t}^{\mathrm{r}}),\mathbf{R}_{i}^{\mathrm{other}}|\theta_{i}^{\mathrm{r},Q})],\nonumber
\end{align}
\begin{align}\label{multi_routing_actor}
&\nabla_{\theta_{i}^{\mathrm{r},\mu}}J_{i}^{\mathrm{r}}\nonumber\\
&\approx\mathbb{E}_{s_{i}^{\mathrm{r}}\sim E}\left[\nabla_{\theta_{i}^{\mathrm{r},\mu}}Q_{i}^{\mathrm{r}}(s_{i}^{\mathrm{r}},\mathbf{a}_{i}^{\mathrm{r}},\mathbf{R}_{i}^{\mathrm{other}}|\theta_{i}^{\mathrm{r},Q})|_{s_{i}^{\mathrm{r}}=s_{i,t}^{\mathrm{r}},\mathbf{a}_{i}^{\mathrm{r}}=\mu_{i}^{\mathrm{r}}(s_{i,t}^{\mathrm{r}}|\theta_{i}^{\mathrm{r},\mu})}\right]\nonumber\\
&=\mathbb{E}_{s_{i}^{\mathrm{r}}\sim E}[\nabla_{\mathbf{a}_{i}^{\mathrm{r}}}Q_{i}^{\mathrm{r}}(s_{i}^{\mathrm{r}},\mathbf{a}_{i}^{\mathrm{r}},\mathbf{R}_{i}^{\mathrm{other}}|\theta_{i}^{\mathrm{r},Q})|_{s_{i}^{\mathrm{r}}=s_{i,t}^{\mathrm{r}},\mathbf{a}_{i}^{\mathrm{r}}=\mu_{i}^{\mathrm{r}}(s_{i,t}^{\mathrm{r}})}\nonumber
\\
&\nabla_{\theta_{i}^{\mathrm{r},\mu}}\mu_{i}^{\mathrm{r}}(s_{i}^{\mathrm{r}}|\theta_{i}^{\mathrm{r},\mu})|_{s_{i}^{\mathrm{r}}=s_{i,t}^{\mathrm{r}}}],
\end{align}
where $\mathbf{R}_{i}^{\mathrm{other}}=\{ \mathbf{R}_{1},\ldots,\mathbf{R}_{i-1},\mathbf{R}_{i+1},\ldots,\mathbf{R}_{M}\} $; $\theta_{i}^{\mathrm{r},Q}$, and $\theta_{i}^{\mathrm{r},Q^{'}}$ and $\theta_{i}^{\mathrm{r},\mu}$ are the parameters of the online RCN, target RCN and online RAN, respectively.
\begin{figure*}[htb]
\centering{}\includegraphics[scale=0.55]{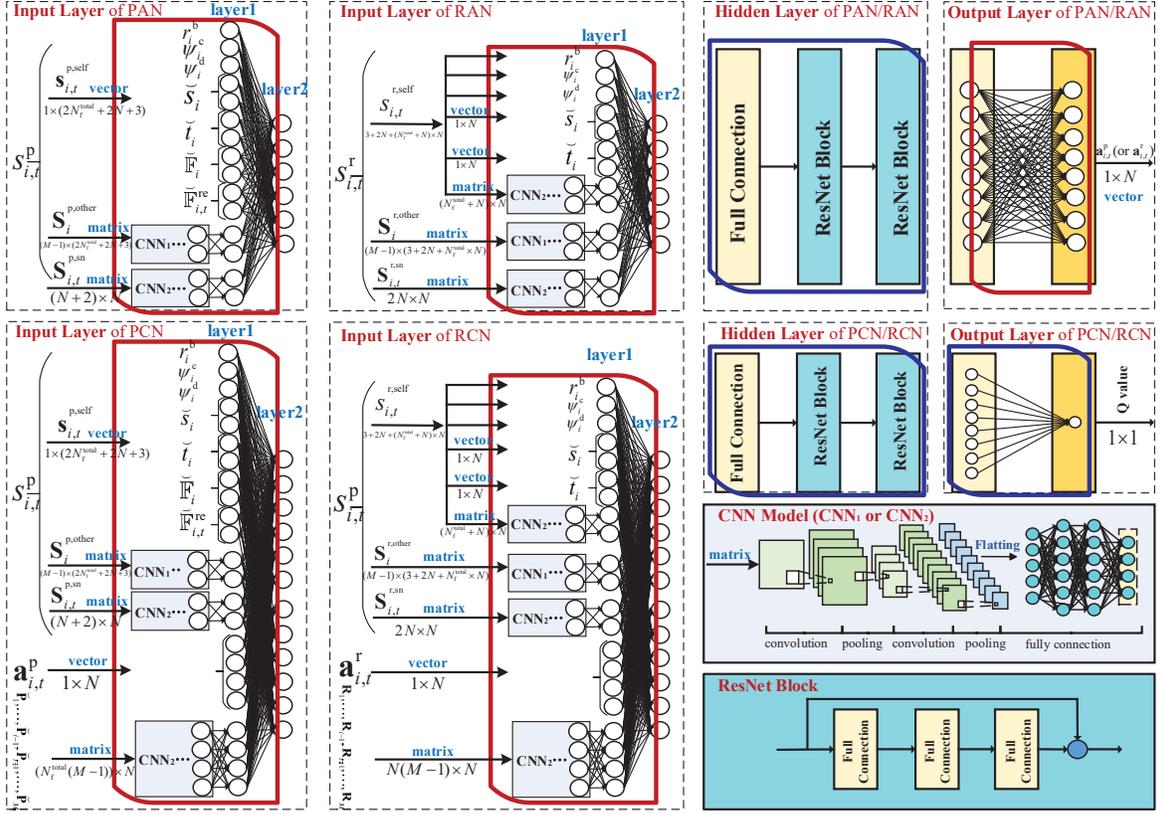}
\caption{The detailed structure of the MADRL-P\&R framework.}\label{placement_stru}
\end{figure*}

Fig. \ref{placement_stru} shows the detailed network structures of the PAN, PCN, RAN and RCN. In $s_{i,t}^{\mathrm{p}}$, the number of elements of $\mathbb{F}_{i}$ and $\mathbb{F}_{i,t}^{\mathrm{re}}$ depends on the variable $N_{\mathrm{f},i}$, which means that the input layer structures of the PAN and PCN are related to the number of VNFs $N_{\mathrm{f},i}$ required by the service $S_{i}$ and is not conducive to the migration of the trained MADRL-P\&R framework to other services. To solve this problem, we map $\mathbb{F}_{i}$ and $\mathbb{F}_{i,t}^{\mathrm{re}}$ to $\mathbb{\breve{F}}_{i}$ and $\mathbb{\breve{F}}_{i,t}^{\mathrm{re}}$, i.e.,
\begin{align}
&(\mathbb{F}_{i})_{1\times N_{\mathrm{f},i}}\rightarrow\mathbb{\breve{F}}_{i}=[\mathbb{\breve{F}}_{i,k}]_{1\times N_{\mathrm{f}}^{\mathrm{total}}},\nonumber \\
&(\mathbb{F}_{i,t}^{\mathrm{re}})_{1\times N_{\mathrm{f},i}}\rightarrow\mathbb{\breve{F}}_{i,t}^{\mathrm{re}}=[\mathbb{\breve{F}}_{i,t,k}^{\mathrm{re}}]_{1\times N_{\mathrm{f}}^{\mathrm{total}}}.
\end{align}
If the $k$-th category VNF ($k\in\left\{ 1,\ldots,N_{\mathrm{f}}^{\mathrm{total}}\right\}$) is required by the $i$-th service, then $\mathbb{\breve{F}}_{i,k}=1$; otherwise, $\mathbb{\breve{F}}_{i,k}=0$. If the $k$-th category VNF required by the $i$-th service completes the placement at step $t$, then $\mathbb{\breve{F}}_{i,t,k}^{\mathrm{re}}\!=\!0$; otherwise, $\mathbb{\breve{F}}_{i,t,k}^{\mathrm{re}}\!=\!1$. Similarly, $\mathbf{P}_{i}$ can also be mapped to
\begin{align}
&\mathbf{P}_{i}=[P_{i,(k,n)}]_{N_{\mathrm{f},i}\times N}\rightarrow\mathbf{\breve{P}}_{i}=[\breve{P}_{i,(k,n)}]_{N_{\mathrm{f}}^{\mathrm{total}}\times N},\nonumber \\
&(s_{i})_{1\times1}\rightarrow(\breve{s}_{i})_{1\times N},(t_{i})_{1\times1}\rightarrow(\breve{t}_{i})_{1\times N}.
\end{align}

Based on $\breve{s}_{i}$, $\breve{t}_{i}$, $\mathbb{\breve{F}}_{i}$, $\mathbb{\breve{F}}_{i,t}^{\mathrm{re}}$ and $\mathbf{\breve{P}}_{i}$, $s_{i,t}^{\mathrm{p}}$ is rewritten as $\ensuremath{s_{i,t}^{\mathrm{p}}=\{ \mathbf{s}_{i,t}^{\mathrm{p,self}},\mathbf{S}_{i}^{\mathrm{p,other}},\mathbf{S}_{i,t}^{\mathrm{p,sn}}\} }$, where $\mathbf{s}_{i,t}^{\mathrm{p,self}}=[r_{i}^{\mathrm{s}},\varphi_{i}^{\mathrm{c}},\varphi_{i}^{\mathrm{d}},\breve{s}_{i},\breve{t}_{i},\mathbb{\breve{F}}_{i},\mathbb{\breve{F}}_{i,t}^{\mathrm{re}}]_{1\times(2N_{\mathrm{f}}^{\mathrm{total}}+2N+3)}$; $\mathbf{S}_{i,t}^{\mathrm{p,sn}}=[[\mathbf{C}_{t}^{\mathrm{re}}]^{\mathrm{T}},[\mathbf{M}_{t}^{\mathrm{re}}]^{\mathrm{T}},\mathbf{A}^{\mathrm{T}}]_{(N+2)\times N}^{\mathrm{T}}$; and $\mathbf{S}_{i}^{\mathrm{p,other}}=[[r_{1}^{\mathrm{s}},\varphi_{1}^{\mathrm{c}},\varphi_{1}^{\mathrm{d}},\breve{s}_{1},\breve{t}_{1},\mathbb{\breve{F}}_{1}]^{\mathrm{T}},\ldots,[r_{M}^{\mathrm{s}},\varphi_{M}^{\mathrm{c}},\varphi_{M}^{\mathrm{d}},\breve{s}_{M},\breve{t}_{M},\mathbb{\breve{F}}_{M}]^{\mathrm{T}}]$. Since $\mathbf{S}_{i,t}^{\mathrm{p,sn}}$ and $\mathbf{S}_{i}^{\mathrm{p,other}}$ are two matrices, two CNNs can be applied, as shown in Fig. \ref{placement_stru}. Here, CNN$_{1}$ and CNN$_{2}$ have different purposes. In addition to facilitating matrix input (i.e., the purpose of CNN$_{1}$), CNN$_{2}$ ia also used to compress the information of other agents. In this way, the proportion of the observation information of the $i$-th agent in all input information is increased. This is conducive to the update of the critic network at the $i$-th agent.

Based on $\mathbf{\breve{P}}_{i}$, $\breve{t}_{i}$ and$\breve{s}_{i}$, $s_{i,t}^{\mathrm{r}}$ can be rewritten as $\ensuremath{s_{i,t}^{\mathrm{r}}=\{ s_{i,t}^{\mathrm{r,self}},\mathbf{S}_{i}^{\mathrm{r,other}},\mathbf{S}_{i,t}^{\mathrm{r,sn}}\} }$, where $s_{i,t}^{\mathrm{r,self}}=\{r_{i}^{\mathrm{s}},\varphi_{i}^{\mathrm{c}},\varphi_{i}^{\mathrm{d}},\breve{s}_{i},\breve{t}_{i},[[\mathbf{\breve{P}}_{i}]^{\mathrm{T}},[\mathbf{R}_{i,t}^{\mathrm{cu}}]^{\mathrm{T}}]_{(N_{\mathrm{f}}^{\mathrm{total}}+N)\times N}^{\mathrm{T}}\}$; $\mathbf{S}_{i,t}^{\mathrm{r,sn}}=[[\mathbf{B}_{t}^{\mathrm{re}}]^{\mathrm{T}},\mathbf{A}^{\mathrm{T}}]_{(2N)\times N}^{\mathrm{T}}$; and  $\mathbf{S}_{i}^{\mathrm{r,other}}
=[[r_{1}^{\mathrm{s}},\varphi_{1}^{\mathrm{c}},\varphi_{1}^{\mathrm{d}},\breve{s}_{1},\breve{t}_{1},\mathbf{v}(\mathbf{\breve{P}}_{1})]^{\mathrm{T}},\ldots,
[r_{M}^{\mathrm{s}},\varphi_{M}^{\mathrm{c}},\varphi_{M}^{\mathrm{d}},\breve{s}_{M},\breve{t}_{M},\mathbf{v}(\mathbf{\breve{P}}_{M})]^{\mathrm{T}}]$. $\mathbf{v}(\mathbf{\breve{P}}_{i})=[\breve{\mathbf{p}}_{i,1},\ldots\breve{\mathbf{p}}_{i,k},\ldots,\breve{\mathbf{p}}_{i,N_{\mathrm{f}}^{\mathrm{total}}}]$, and $\breve{\mathbf{p}}_{i,k}$ is the $k$-th row of matrix $\mathbf{\breve{P}}_{i}$.

Due to the complexity of \emph{OP1}, this paper uses a large number of network layers. However, the large number of network layers also brings the gradient dispersion. Considering that the residual network (ResNet) can solve this challenge, part of fully connected neural network (FCNN) in this paper is replaced by the ResNet, as shown in Fig. \ref{placement_stru}.
\subsection{Model Training and Parameter Migration}
Algorithms 1 and 2 summarize the training process\footnote{In our multi-agent algorithm, the public information includes: 1) Network topology and resource information reflected in state observations $s^{\mathrm{p}_{i,t}}$ and $s^{\mathrm{r}_{i,t}}$ can be observed by all agents simultaneously (the third line in Algorithm 1); and 2) Other agent action information required by the centralized action-value function. Due to the parallel operation of all agents, this information can be updated uniformly after all the agents run (lines 5--12 and lines 22--32 in Algorithm 2).}, where $EM_{i}^{\mathrm{p}}$ and $EM_{i}^{\mathrm{r}}$ respectively stand for the experience replay memories of the $i$-th placement and routing agents, respectively. $N_{\mathrm{ep}}$ is the number of epochs (defined in Section V-A), and $N_{\mathrm{max}}^{\mathrm{p}}$ and $N_{\mathrm{max}}^{\mathrm{r}}$ denote the maximum number of episodes for placement and routing agents to explore SPRs and SRRs in each epoch. Specifically, for avoiding the problem that the agent learning cannot be terminated due to C16--C18 when remaining resources are lower than that required, we define the $N_{\mathrm{p}}^{\mathrm{pre}}$ and $N_{\mathrm{r}}^{\mathrm{pre}}$ to limit the number of epoch for the placement and routing modules to explore SPRs and SRRs.
\begin{algorithm}[htbp]
\footnotesize
\caption{Training the MADRL-P\&R}
\label{algorithm1}
\setstretch{0.4}
\LinesNumbered
\SetKwInOut{KwIn}{\textbf{Initialize}}
\KwIn{$\theta_{i}^{\mathrm{p},\mu}$, $\theta_{i}^{\mathrm{p},\mu^{'}}$, $\theta_{i}^{\mathrm{p},Q}$, $\theta_{i}^{\mathrm{p},Q^{'}}$, $\theta_{i}^{\mathrm{r},\mu}$, $\theta_{i}^{\mathrm{r},\mu^{'}}$, $\theta_{i}^{\mathrm{r},Q}$, $\theta_{i}^{\mathrm{r},Q^{'}}$, $N_{\mathrm{p}}=0$, $N_{\mathrm{r}}=0$}
\For{Epoch = 1,2,...,$N_{\mathrm{ep}}$}{
Initialize two random processes to explore the placement action and routing action\;
Receive initial states $s_{i,t}^{\mathrm{p}}$ and $s_{i,t}^{\mathrm{r}}$ for all agents\;
Execute PAN with a total of $N_{\mathrm{f},i}$ steps and record $s_{i,t}^{\mathrm{p}}$, $s_{i,t+1}^{\mathrm{p}}$ and $\mathbf{a}_{i,t}^{\mathrm{p}}$ of each decision step for all placement agents\;
Get placement results $\left\{ \mathbf{P}_{1},\mathbf{P}_{2},\ldots,\mathbf{P}_{M}\right\}$ by (21)\;
\eIf{$\left\{ \mathbf{P}_{1},\mathbf{P}_{2},\ldots,\mathbf{P}_{M}\right\}$ are all SPRs or $N_{\mathrm{p}}\geq N_{\mathrm{p}}^{\mathrm{pre}}$}{
Execute RAN with a total of $N-1$ steps, record $s_{i,t}^{\mathrm{r}}$, $s_{i,t+1}^{\mathrm{r}}$ and $\mathbf{a}_{i,t}^{\mathrm{r}}$ of each decision step for all routing agents, and get routing results $\left\{ \mathbf{R}_{1},\mathbf{R}_{2},\ldots,\mathbf{R}_{M}\right\}$ by (24)\;
\eIf{$\left\{ \mathbf{R}_{1},\mathbf{R}_{2},\ldots,\mathbf{R}_{M}\right\} $ are all SRRs or $N_{\mathrm{r}}\geq N_{\mathrm{r}}^{\mathrm{pre}}$}{
Calculate \emph{OP1}, obtain joint reward $\ensuremath{r_{i,t}^{\mathrm{p,jo}}}$ and $\ensuremath{r_{i,t}^{\mathrm{r,jo}}}$ for each placement and routing decision steps of all agents by (27)\;
Calculate final reward $r_{i,t}^{\mathrm{p,fin}}$ and $r_{i,t}^{\mathrm{r,fin}}$\;
Store $(s_{i,t}^{\mathrm{p}},\mathbf{a}_{i,t}^{\mathrm{p}},\mathbf{P}_{i}^{\mathrm{other}},r_{i,t}^{\mathrm{p,fin}},s_{i,t+1}^{\mathrm{p}})$ of each placement decision step to $EM_{i}^{\mathrm{p}}$ for all placement agents, and store $(s_{i,t}^{\mathrm{r}},\mathbf{a}_{i,t}^{\mathrm{r}},\mathbf{R}_{i}^{\mathrm{other}},r_{i,t}^{\mathrm{r,fin}},s_{i,t+1}^{\mathrm{r}})$ of each routing decision step to $EM_{i}^{\mathrm{r}}$ for all routing agents\;
\If{all $EM_{i}^{\mathrm{p}}$ are full}{\textbf{Procedure 1}: \emph{\textbf{Training placement agent}}}
\If{all $EM_{i}^{\mathrm{r}}$ are full}{\textbf{Procedure 2}: \emph{\textbf{Training routing agent}} in Algorithm 2}
}{Run \textbf{placement module} in Algorithm 2}
}{Run \textbf{routing module} in Algorithm 2}
}
\BlankLine
\textbf{Procedure 1}: \emph{\textbf{Training placement module}}\;
\For{placement agent $i$ = 1,...,$M$}{
Randomly sample mini-batch from $EM_{i}^{\mathrm{p}}$\;
Update PCN and PAN of the $i$-th placement agent by (30) and (31), and ``Soft update'' target networks\;
}
\end{algorithm}

\begin{algorithm}[htbp]
\footnotesize
\caption{Network update algorithms}
\label{algorithm2}
\setstretch{0.5}
\LinesNumbered
Run \textbf{placement module}:\;
$N_{\mathrm{p}}=N_{\mathrm{p}}+1$\;
\For{Episode = 1,2,...,$N^{\mathrm{max}}_{\mathrm{p}}$}{
\For{placement agent $i=1,...,M$}{
\For{$t$ = 1,2,...,$N_{\mathrm{f,max}}$}{
Input $s_{i,t}^{\mathrm{p}}$ to PAN and output $\mathbf{a}_{i,t}^{\mathrm{p}}$\;
Acquire internal reward $r_{i,t}^{\mathrm{p,int}}$ and calculate $r_{i,t}^{\mathrm{p,fin}}$\;
Observe the next placement state $s_{i,t+1}^{\mathrm{p}}$\;
Record $s_{i,t}^{\mathrm{p}}$, $\mathbf{a}_{i,t}^{\mathrm{p}}$, and $r_{i,t}^{\mathrm{p,fin}}$, $s_{i,t+1}^{\mathrm{p}}$\;
}}
Get placement results $\left\{ \mathbf{P}_{1},\mathbf{P}_{2},\ldots,\mathbf{P}_{M}\right\}$ by (21)\;
Store $(s_{i,t}^{\mathrm{p}},\mathbf{a}_{i,t}^{\mathrm{p}},\mathbf{P}_{i}^{\mathrm{other}},r_{i,t}^{\mathrm{p,fin}},s_{i,t+1}^{\mathrm{p}})$ of each placement decision step to $EM_{i}^{\mathrm{p}}$ for all agents\;
\If{all $EM_{i}^{\mathrm{p}}$ are full}{
\textbf{Procedure 1}: \emph{\textbf{Training placement agent}}\;
}}
\BlankLine
\BlankLine
Run \textbf{routing module}:\;
$N_{\mathrm{r}}=N_{\mathrm{r}}+1$\;
\For{Episode = 1,2,...,$N^{\mathrm{max}}_{\mathrm{r}}$}{
\For{routing agent $i=1,...,M$}{
\For{$t$ = 1,2,...,$N-1$}{
Input $s_{i,t}^{\mathrm{r}}$ to RAN and output $\mathbf{a}_{i,t}^{\mathrm{r}}$\;
Acquire internal reward $r_{i,t}^{\mathrm{r,int}}$ and calculate $r_{i,t}^{\mathrm{r,fin}}$\;
Observe the next routing state $s_{i,t+1}^{\mathrm{r}}$\;
Record $s_{i,t}^{\mathrm{r}}$, $\mathbf{a}_{i,t}^{\mathrm{r}}$, and $r_{i,t}^{\mathrm{r,fin}}$, $s_{i,t+1}^{\mathrm{r}}$\;
\If{$\mathbf{a}_{i,t}^{\mathrm{r}}$ is the destination node}{Break\;}
}}
Get routing results $\left\{ \mathbf{R}_{1},\mathbf{R}_{2},\ldots,\mathbf{R}_{M}\right\}$ by (24)\;
Store $(s_{i,t}^{\mathrm{r}},\mathbf{a}_{i,t}^{\mathrm{r}},\mathbf{R}_{i}^{\mathrm{other}},r_{i,t}^{\mathrm{r,fin}},s_{i,t+1}^{\mathrm{r}})$ of each routing decision step to $EM_{i}^{\mathrm{r}}$ for all agents\;
\If{all $EM_{i}^{\mathrm{r}}$ are full}{\textbf{Procedure 2}: \emph{\textbf{Training routing module}}\;}
}
\BlankLine
\BlankLine
\textbf{Procedure 2}: \emph{\textbf{Training routing module}}\;
\For{routing agent $i$ = 1,...,$M$}{
Randomly sample mini-batch from $EM_{i}^{\mathrm{r}}$\;
Update RCN and RAN of the $i$-th placement agent by (32) and (33), and ``Soft update'' target networks\;
}
\end{algorithm}

In this case, our MADRL-P\&R pursues the deployment of the largest possible number of service requests. To this end, we also adopted a batch-by-batch service request deployment mechanism: build a MADRL framework consisting of $2M_{\mathrm{batch}}$ ($M_{\mathrm{batch}}<M$) agents, and output a batch ($M_{\mathrm{batch}}$ requests in each batch) of service request deployment policies at a time until there are no resources available for deployment\footnote{To ensure the $M$ is exactly divisible by $M_{\mathrm{batch}}$, additional service requests with a demand of 0 need to be supplemented.}. Moreover, the batch-by-batch deployment also avoids the problem that $M$ agents cannot be constructed simultaneously due to the limited of CPU resources and storage space, when $M$ is large.

A trained RL model usually performs well in a specific environment. As a result, we need to rebuild and retrain the MADRL-P\&R model when the network topology (including the nodes or links) changes. This is unacceptable in practice, due to the cost of retraining. Considering that a trained RL model already has a certain ability to place and route based on constraints and the optimization objective, we propose a ``parameter migration'' method, which divides all PANs, PCNs, RANs and RCNs into two parts: the reconfigurable and fixed networks. The reconfigurable network refers to the part of the network where the \emph{number of neurons} and \emph{network parameters} need to be adjusted and retrained when the network topology changes, as shown in the red boxes of Fig. \ref{placement_stru}. All the rest of the network are fixed networks and remain unchanged based on the ``parameter freezing'' technique\footnote{Parameter freezing is the key to transfer learning \cite{weiss2016survey}, and can be easily implemented (including the back-propagation) in existing mainstream learning architectures, e.g., TensorFlow, Pytorch.}, as shown in the blue boxes of Fig. \ref{placement_stru}. When only the network links change (the number of nodes remains unchanged), we only retrain the network parameters in the reconfigurable part without adjusting the number of neurons.
\subsection{Time Complexity Analysis}
Let $N_{\mathrm{full}}$ be the number of neurons of FCNN, $d_{\mathrm{cnn}}$ be the number of the flattened layer in CNN$_{1}$ and CNN$_{2}$, $d_{\mathrm{Res}}$ be the number of ResNet blocks ($N_{\mathrm{full}}\!>\!N\!\gg\! N_{\mathrm{f}}^{\mathrm{total}}\!>\!d_{\mathrm{cnn}}\!=\!d_{\mathrm{Res}}$), $N_{\mathrm{b}}$ be the batch size, $D$ be the total number of the convolutional layers, $l$ be the $l$-th convolutional layer, $C_{l}$ be the number of convolution channels, $K_{l}$ and $M_{l}$ be respectively the side length of the convolution kernel and the side length of the output feature map.
The training time for each step consists of the feed-forward propagation time and the back-propagation time, and both have the same complexity \cite{286919}. In each step, the time complexity of feed-forward propagation and back-propagation are both
$\mathcal{O}(N_{\mathrm{b}}(\sum_{l}(M_{l}^{2}K_{l}^{2}C_{l-1}C_{l})+(d_{\mathrm{net}}N_{\mathrm{full}}+7N)N_{\mathrm{full}}))$ for PAN ($d_{\mathrm{net}}=d_{\mathrm{cnn}}+3d_{\mathrm{Res}}+1$); and the time complexity of feed-forward propagation and back-propagation are both
$\mathcal{O}(N_{\mathrm{b}}(\sum_{l}(M_{l}^{2}K_{l}^{2}C_{l-1}C_{l})+(d_{\mathrm{net}}N_{\mathrm{full}}+10N)N_{\mathrm{full}}))$ for PCN. Similarly, the time complexity of feed-forward propagation (or back-propagation) of RAN and RCN are $\ensuremath{\mathcal{O}(N_{\mathrm{b}}(\sum_{l}(M_{l}^{2}K_{l}^{2}C_{l-1}C_{l})+(d_{\mathrm{net}}N_{\mathrm{full}}+11N)N_{\mathrm{full}}))}$ and $\mathcal{O}(N_{\mathrm{b}}(\sum_{l}(M_{l}^{2}K_{l}^{2}C_{l-1}C_{l})+(d_{\mathrm{net}}N_{\mathrm{full}}+13N)N_{\mathrm{full}}))$, respectively. In Algorithm 1, a back-propagation is performed after each execution of at most $N_{\mathrm{f}}^{\mathrm{max}}$ decision steps of the placement module or after every execution of at most $N-1$ decision steps of the routing module. Thus, the total training time of  the $i$-th placement agent and routing agent is
$\mathcal{O}(2N_{\mathrm{b}}N_{\mathrm{ep}}((N_{\mathrm{max}}^{\mathrm{p}}N_{\mathrm{f}}^{\mathrm{max}}+N_{\mathrm{max}}^{\mathrm{r}}N)(\sum_{l}(M_{l}^{2}K_{l}^{2}C_{l-1}C_{l})+d_{\mathrm{net}}N_{\mathrm{full}})+(17N_{\mathrm{max}}^{\mathrm{p}}N_{\mathrm{f}}^{\mathrm{max}}+24N_{\mathrm{max}}^{\mathrm{r}}N)NN_{\mathrm{full}}))$. Note that $M$ agents in the placement and routing modules have the same structure and operate in parallel, the total training time-complexity of our framework is the same as that for a single agent. For the trained MADRL-P\&R model, the time complexity (i.e., the inference time) of service request deployment depends only on the feed-forward propagation time, so it can be expressed as $\mathcal{O}((N_{\mathrm{f}}^{\mathrm{max}}+N)(\sum_{l}(M_{l}^{2}K_{l}^{2}C_{l-1}C_{l})+d_{\mathrm{net}}N_{\mathrm{full}}^{2})+(17N_{\mathrm{f}}^{\mathrm{max}}+24N)NN_{\mathrm{full}})$.
\section{Performance Evaluation}
The proposed MADRL-P\&R framework is evaluated in this section. For comparison purpose, five state-of-the-art alternatives are implemented:
\begin{itemize}
\item DL-TPA \cite{9062539}. It first solves the binary integer programming with a near-optimal algorithm to generate training data \cite{8058433}. Then, the VNF chaining (or routing) network and VNF selection network are designed and trained for solving the VNF routing and placement in two phases. In DL-TPA, the SFCs are mapped into the SN one by one, until all SFCs are served or resources are exhausted, instead of deploying multiple SFCs simultaneously.
\item PG \cite{li2019virtual}. It first classifies the SFCs into different clusters based on the source nodes, and each cluster and each SFC in each cluster are prioritized based on the number of VNFs in SFCs. Then, the SFCs are mapped into the SN one cluster after another. The routing path of an SFC is obtained using the shortest path algorithm.
\item HMFMR \cite{8463518}. It first iterates all required VNFs for each service, and places the VNF with the highest resource demand on the node with the largest remaining capacity in turn, according to the best-fit decreasing method \cite{man1996approximation}. Then, the routing paths are established following the shortest path algorithm (regarding link delays).
\item KPMST-H \cite{8648029}. It first builds a key-node preferred minimum spanning tree (KPMST)-based routing tree, then places VNFs in a greedy fashion.
\item JVPR \cite{9468711}. It constructs a generic heuristic optimization structure called as VNF-splited multi-stage edge-weight graph, and devises a heuristic algorithm to solve.  To make JVPR comparable with the proposed MADRL-P\&R, the delay needs to be added into (33) in \cite{9468711}.
\end{itemize}

When $M$ is small ($M\leq400$), we can also calculate the optimal solutions by employing the Lingo solver with branch-and-bound method, where C9 needs to be converted to a linear constraint, as done in [6, eqs. 20\&21].
\subsection{Simulation Setup}
The simulated network topology includes: the ``COST266'' and ``TA2'' topologies from SDNlib, and the two larger network topologies generated by ourselves referring to the method in \cite{8648029}\footnote{The COST266 is composed of 37 nodes and 57 links; the TA2 composed of 65 nodes and 108 links; the first network topology we generated ourselves (denoted as ``SELF1'') is composed of 100 nodes and 324 links; and the second network topology we generated ourselves (denoted as ``SELF2'') is composed of 225 nodes and 784 links.}. The cost per unit of computing resources $c_{n}^{\mathrm{c}}$ at node $n$ is uniformly randomly distributed from 1 to 3, i.e., $c_{n}^{\mathrm{c}}\sim\sim\mathcal{U}\left(1,3\right)$. The other parameters related to the cost and delay are set as follows: $c_{n}^{\mathrm{m}}\sim\mathcal{U}\left(1,3\right)$, $c_{\left(u,v\right)}^{\mathrm{B}}\sim\mathcal{U}\left(5,15\right)$, $c_{k,n}^{\mathrm{d}}\sim\mathcal{U}\left(5,15\right)$, $\eta_{k}\sim\mathcal{U}\left(0.2,1\right)$, $f_{k}^{\mathrm{m}}\sim\mathcal{U}\left(1,5\right)$, $C_{n}\sim\mathcal{U}\left(250,350\right)$, $M_{n}\sim\mathcal{U}\left(250,350\right)$, $d_{k}^{\mathrm{dyn}}\sim\mathcal{U}\left(0.5,3\right)$, $d_{\left(u,v\right)}^{\mathrm{B}}\sim\mathcal{U}\left(0.5,3\right)$ and $d^{\mathrm{inv}}=1$. The number of VNFs required for each service request satisfies: $N_{\mathrm{f},i}\in\left\{ 2,3,4\right\}$. The number of VNF categories is 10. The detailed structure of the MADRL-P\&R framework is as shown in Fig. \ref{placement_stru}, where each ``Full Connection'' layer has 64 neurons and adopts the ReLU activation function. The activation function of the output layer in the PAN and RAN is Softmax. The mini-batch gradient descent is conducted by the RMSProp algorithm, and the bath size is 256. Both $EM_{i}^{\mathrm{p}}$ and $EM_{i}^{\mathrm{r}}$ are set to 4,000. The learning rates of  the PAN and RAN are 0.002 and 0.001, respectively. The learning rates of  the PCN and RCN are 0.05 and 0.01, respectively. The other hyper-parameters are: $\omega_{i}^{\mathrm{p,int}}=-10$, $\omega_{\mathrm{I},i}^{\mathrm{r,int}}=-8$, $\omega_{\mathrm{II},i}^{\mathrm{r,int}}=-2$, $\omega_{\mathrm{III},i}^{\mathrm{r,int}}=-1$, $\omega_{\mathrm{scal}}^{\mathrm{jo,exp}}=-\left(10M\right)^{-1}$, $\omega_{\mathrm{trans}}^{\mathrm{jo,exp}}=20$, $\gamma=0.99$, $\psi^{\mathrm{c}}=\psi^{\mathrm{d}}=1$, $N_{\mathrm{ep}}=10,000$, $N_{\mathrm{max}}^{\mathrm{p}}=500$, $N_{\mathrm{max}}^{\mathrm{r}}=1,000$ and $N_{\mathrm{p}}^{\mathrm{pre}}=N_{\mathrm{r}}^{\mathrm{pre}}=4,500$.
\subsection{Comparison of Cost and Delay}
\begin{figure}[h]
  \centering
  \subfigure[$r_{i}^{\mathrm{b}}$=5.4 Mbps, $M\!=\!400$, $\overline{\varphi_{i}^{\mathrm{d}}}\!=\!0.5$]{
    \includegraphics[width=3.1in]{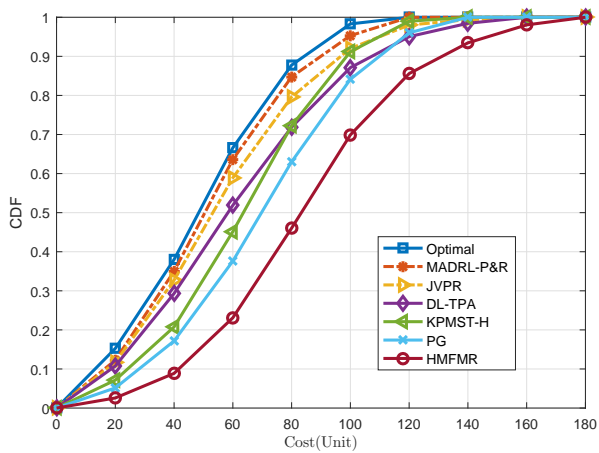}
  }
  \subfigure[$r_{i}^{\mathrm{b}}$=5.4 Mbps, $M\!=\!400$, $\overline{\varphi_{i}^{\mathrm{d}}}\!=\!0.5$]{
    \includegraphics[width=3.1in]{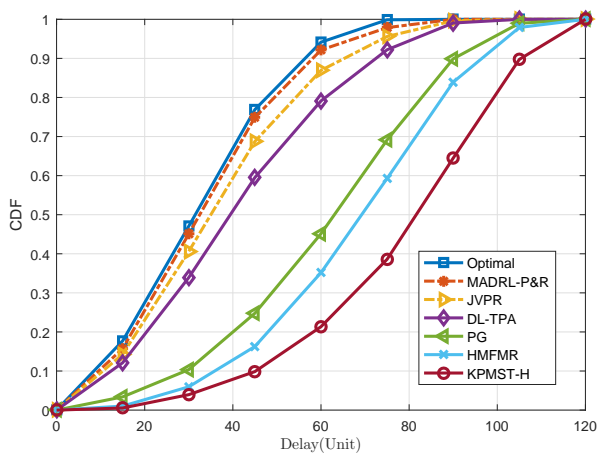}
  }
  \caption{COST266 network topology. (a): CDF of cost; and (b): CDF of delay.}
  \label{cost_delay}
\end{figure}
Figs. \ref{cost_delay}(a) and (b) show the cumulative distribution function (CDF) of cost and delay on the COST266 network topology when the average delay-sensitive factor ($\overline{\varphi_{i}^{\mathrm{d}}}=\left(\sum_{i}\varphi_{i}^{\mathrm{d}}\right)/M$) is 0.5 and the number of service requests is 400. The proposed MADRL-P\&R algorithm is superior to the other algorithms and is close-to-optimal, which means that our method can make the most effective use of network resources. For cost, KPMST-H (average cost is 72.90) is close to DL-TPA (average cost is 71.11), but the delay of KPMST-H (average delay is 85.73) is much longer than DL-TPA (average delay is 48.61). This is because the focus of KPMST-H is on cost optimization, while the focuses of DL-TPA, PG and HMFMR are on the delay optimization. Compared with DL-TPA, PG, HMFMR and HPMST-H, JVPR considers both cost and delay optimization, thus it performs better in terms of cost and delay.

\begin{figure}[h]
  \centering
  \subfigure[$r_{i}^{\mathrm{b}}$=5.4 Mbps, $M\!=\!400$, $\overline{\varphi_{i}^{\mathrm{d}}}\!=\!0.5$]{
    \includegraphics[width=3.1in]{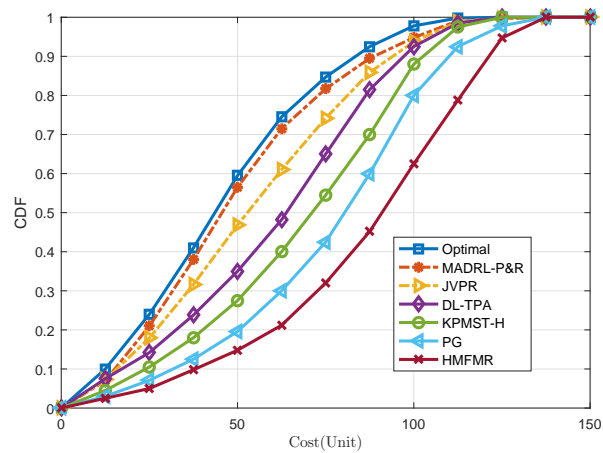}
  }
  \subfigure[$r_{i}^{\mathrm{b}}$=5.4 Mbps, $M\!=\!400$, $\overline{\varphi_{i}^{\mathrm{d}}}\!=\!0.5$]{
    \includegraphics[width=3.1in]{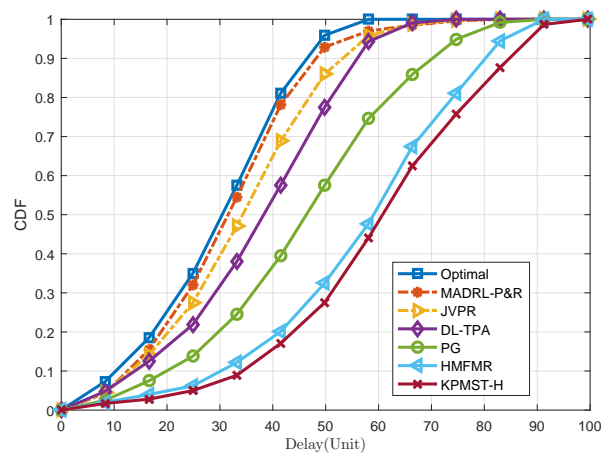}
  }
  \caption{TA2 network topology. (a): CDF of cost; (b): CDF of delay.}
  \label{TA2_R3}
\end{figure}
\begin{figure}[h]
  \centering
  \subfigure[$r_{i}^{\mathrm{b}}$=5.4 Mbps, $M\!=\!400$, $\overline{\varphi_{i}^{\mathrm{d}}}\!=\!0.5$]{
    \includegraphics[width=3.1in]{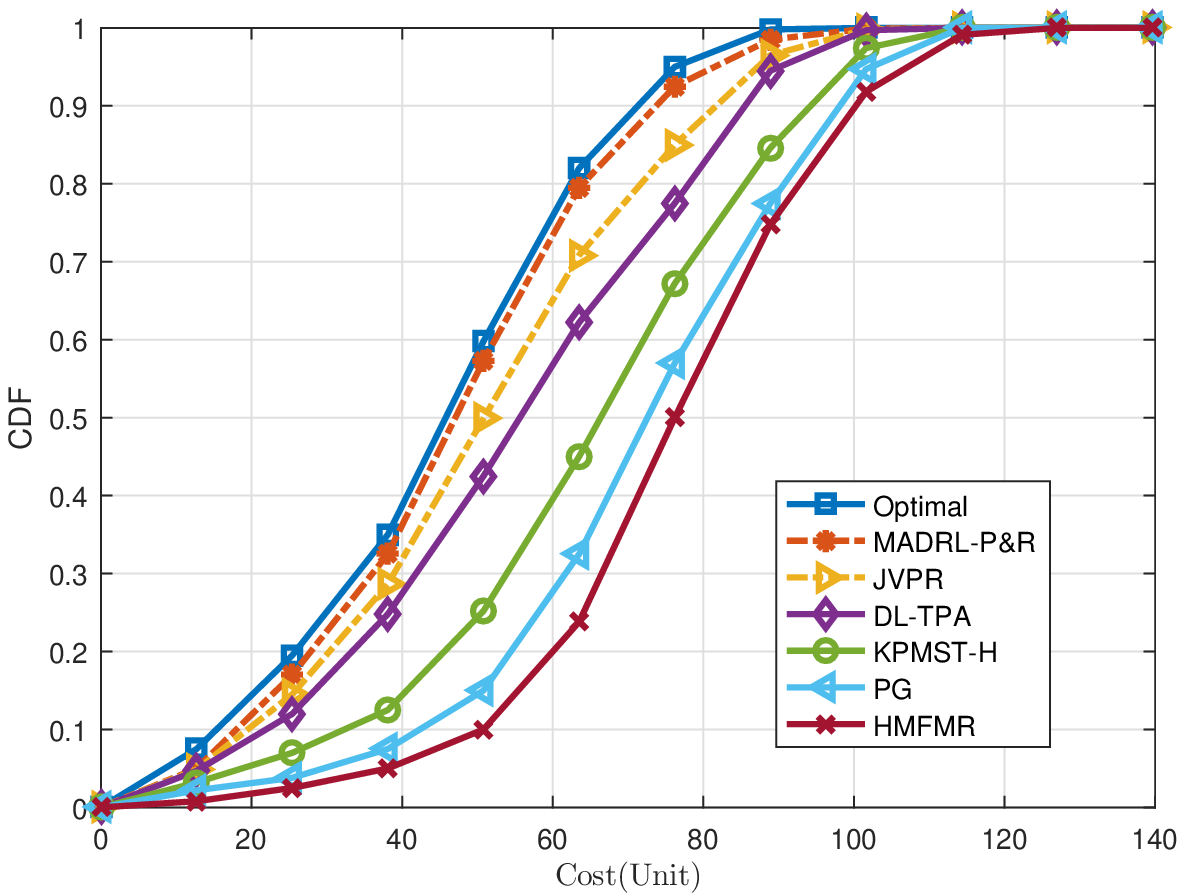}
  }
  \subfigure[$r_{i}^{\mathrm{b}}$=5.4 Mbps, $M\!=\!400$, $\overline{\varphi_{i}^{\mathrm{d}}}\!=\!0.5$]{
    \includegraphics[width=3.1in]{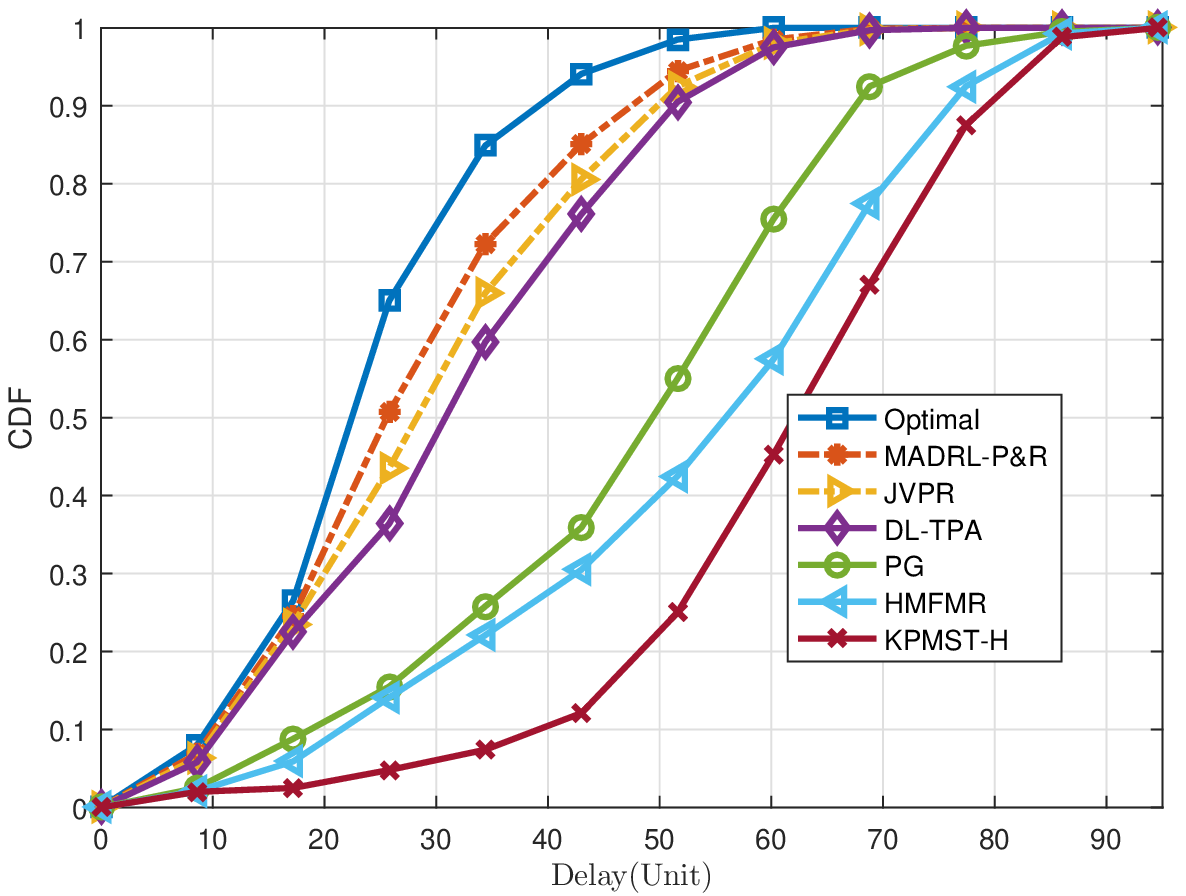}
  }
  \caption{SELF1 network topology. (a): CDF of cost; (b): CDF of delay.}
  \label{our1_R3}
\end{figure}
\begin{figure}[h]
  \centering
  \subfigure[$r_{i}^{\mathrm{b}}$=5.4 Mbps, $M\!=\!400$, $\overline{\varphi_{i}^{\mathrm{d}}}\!=\!0.5$]{
    \includegraphics[width=3.1in]{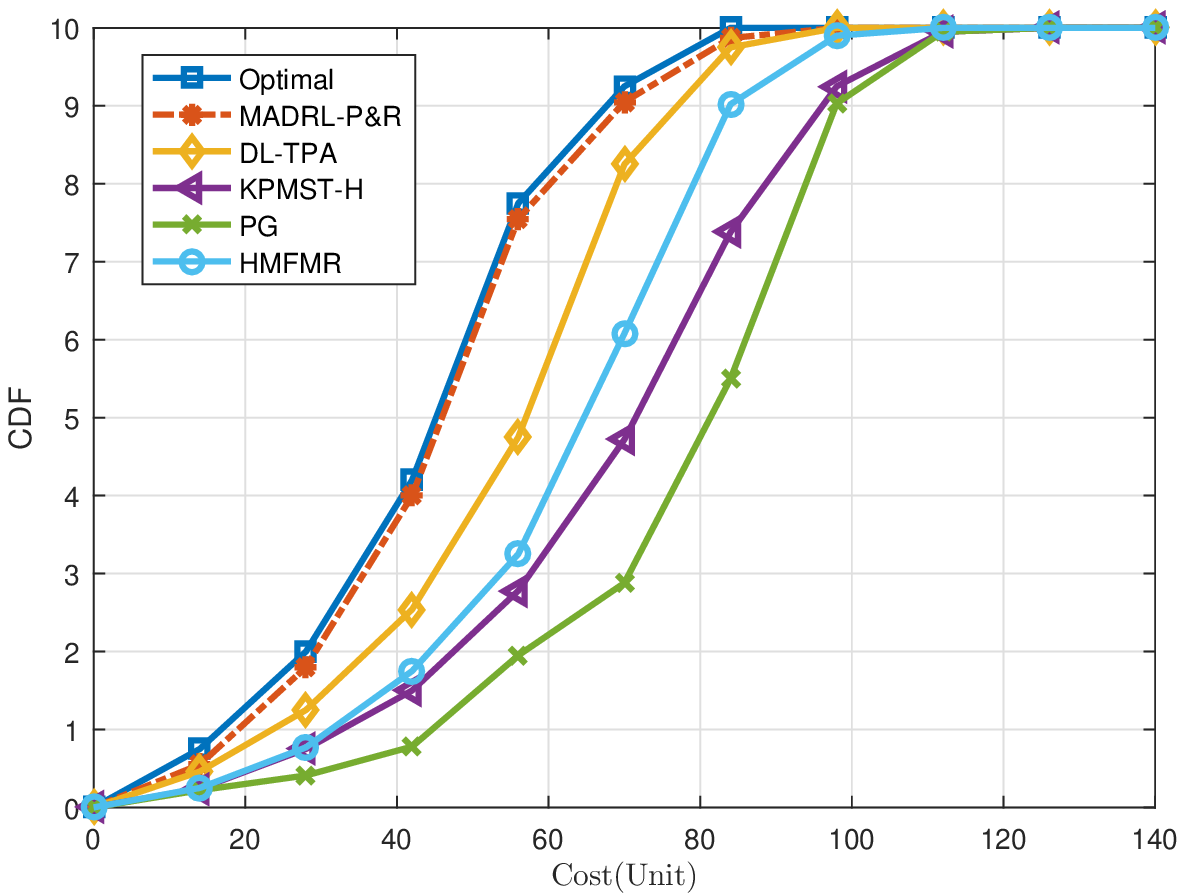}
  }
  \subfigure[$r_{i}^{\mathrm{b}}$=5.4 Mbps, $M\!=\!400$, $\overline{\varphi_{i}^{\mathrm{d}}}\!=\!0.5$]{
    \includegraphics[width=3.1in]{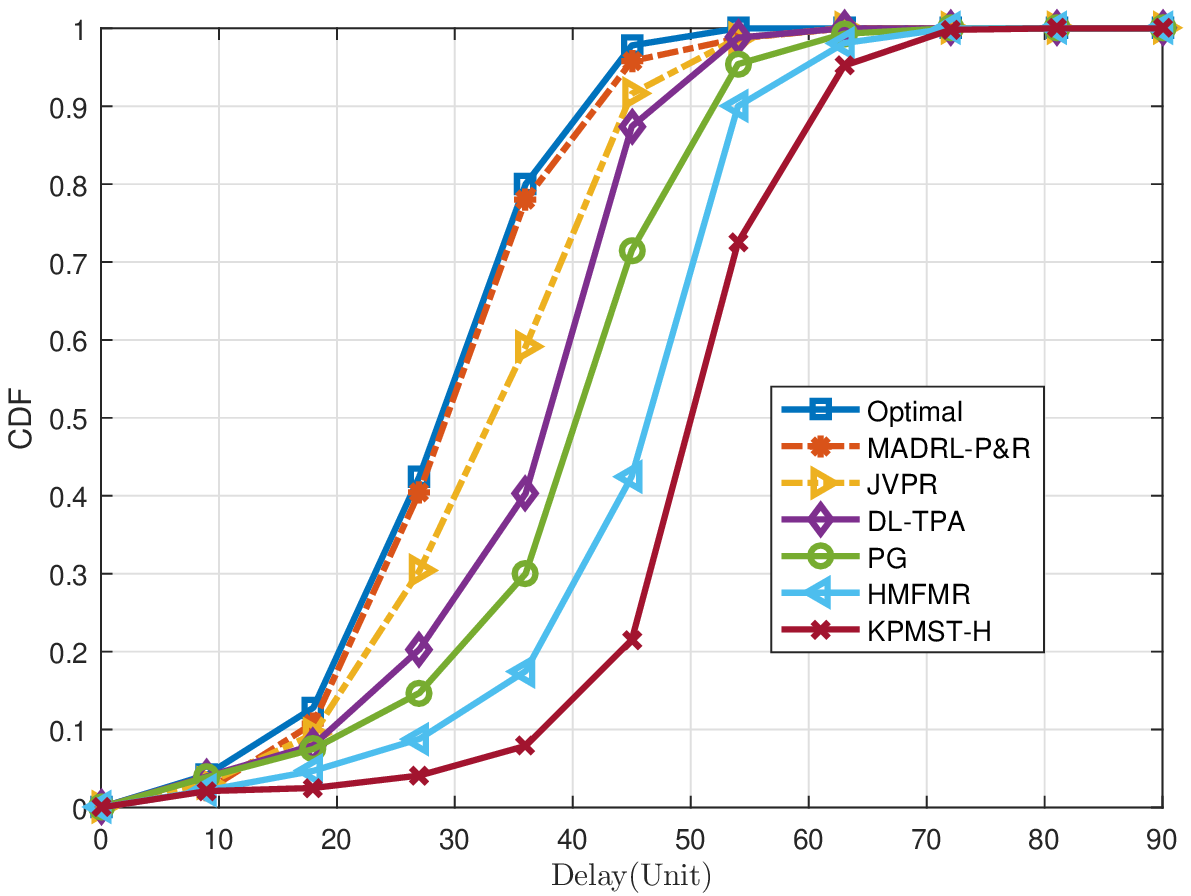}
  }
  \caption{SELF2 network topology. (a): CDF of cost; (b): CDF of delay.}
  \label{our2_R3}
\end{figure}
Figs. \ref{TA2_R3}, \ref{our1_R3} and \ref{our2_R3} shows the CDFs of cost and delay on the TA2, SELF1 and SELF2 network topologies, respectively. Compared to other comparison algorithms, our proposed method performs the best.

\begin{figure}[h]
\centering
\includegraphics[scale=0.65]{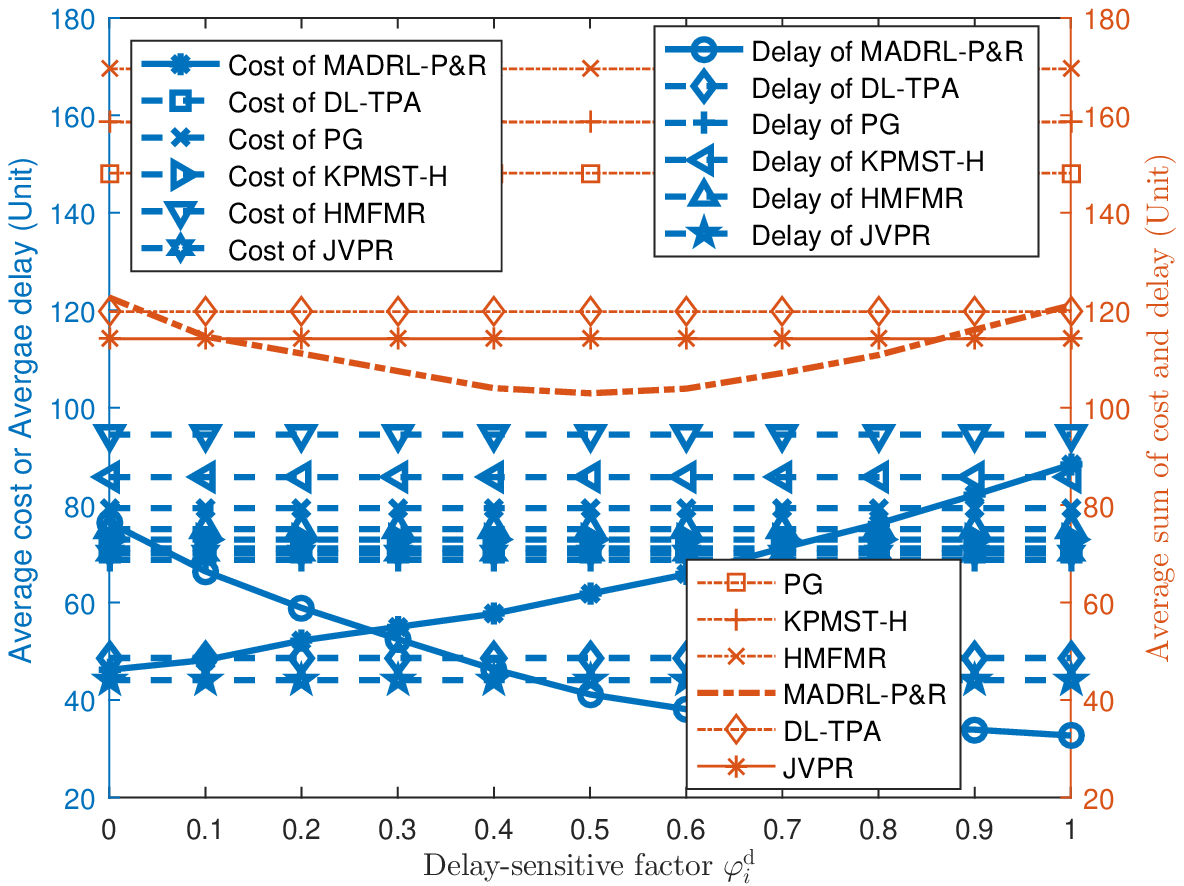}
\caption{The changes in delay-sensitive factor (all $\varphi_{i}^{\mathrm{d}}$ are the same), $r_{i}^{\mathrm{b}}$=5.4 Mbps, $M\!=\!400$, $\overline{\varphi_{i}^{\mathrm{d}}}\!=\!0.5$.}
\label{sensitive_new}
\end{figure}
Fig. \ref{sensitive_new} compares the impact of the delay-sensitive factor on different methods (the delay-sensitive factors of all services are the same), where the cost and delay are the average of each service request. Different from JVPR, DL-TPA, PG, KPMST-H and HMFMR, the new MADRL-P\&R algorithm takes into account the individual needs of different users for delay and cost. Therefore, MADRL-P\&R presents different results with different delay-sensitive factors. This confirms the effectiveness of the MADRL-P\&R method in its support of personalized service requests.
\subsection{Comparison of Throughput and Service Acceptance Ratio}
\begin{figure}[h]
  \centering
  \subfigure[$r_{i}^{\mathrm{b}}$=5.4 Mbps, $\overline{\varphi_{i}^{\mathrm{d}}}=0.5$]{
    \includegraphics[width=3.1in]{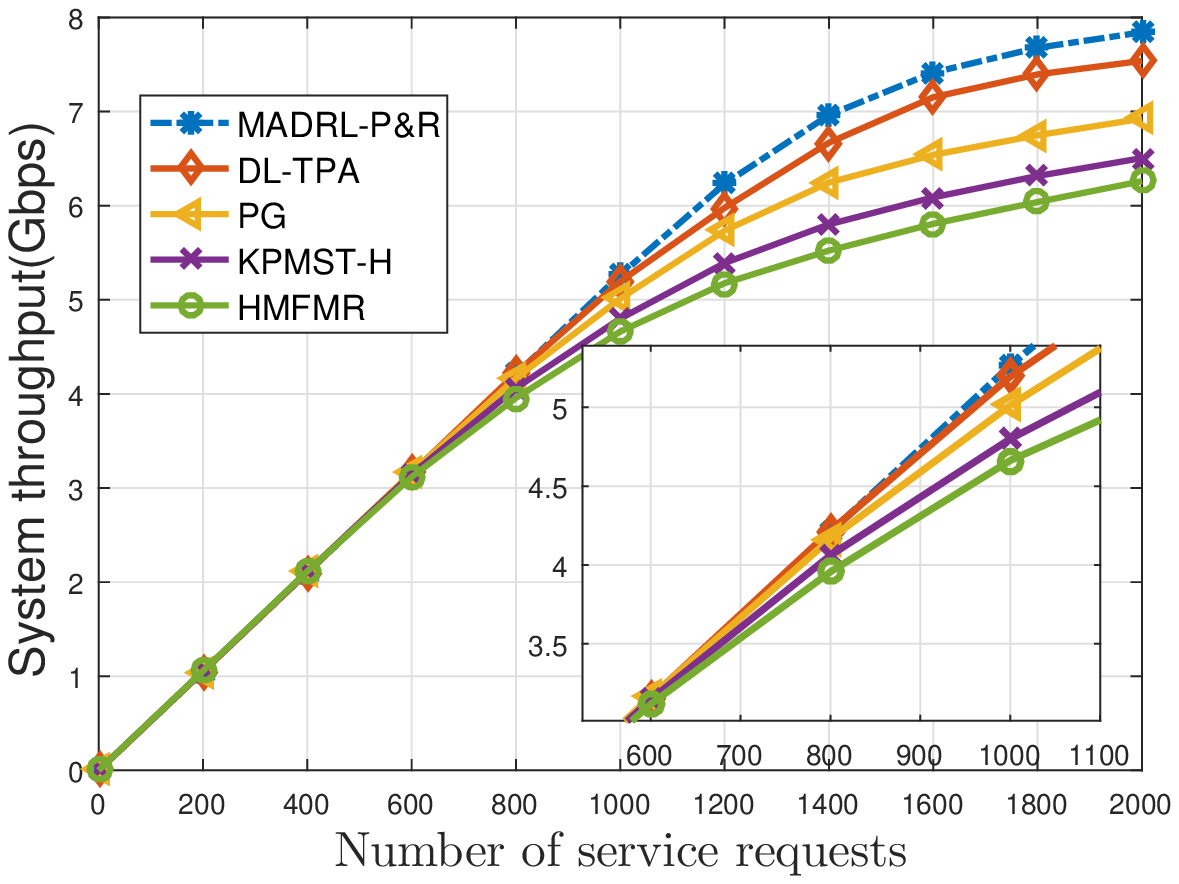}
  }
  \subfigure[$r_{i}^{\mathrm{b}}$=5.4 Mbps, $\overline{\varphi_{i}^{\mathrm{d}}}=0.5$]{
    \includegraphics[width=3.1in]{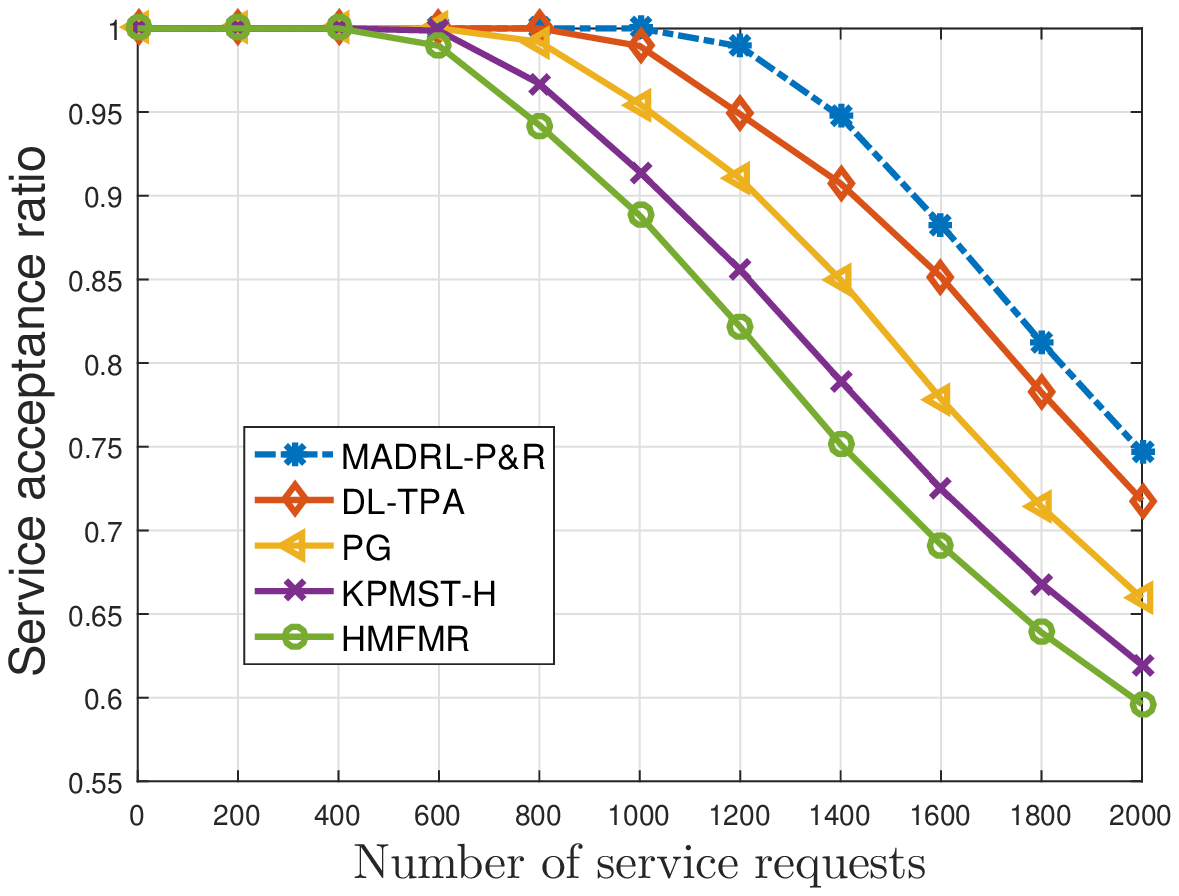}
  }
  \caption{COST266 network topology. (a): Comparison of system throughput; and (b): comparison of service acceptance ratio.}
  \label{throughput_sar}
\end{figure}
\begin{figure}[h]
  \centering
  \subfigure[$r_{i}^{\mathrm{b}}$=5.4 Mbps, $\overline{\varphi_{i}^{\mathrm{d}}}=0.5$]{
    \includegraphics[width=3.1in]{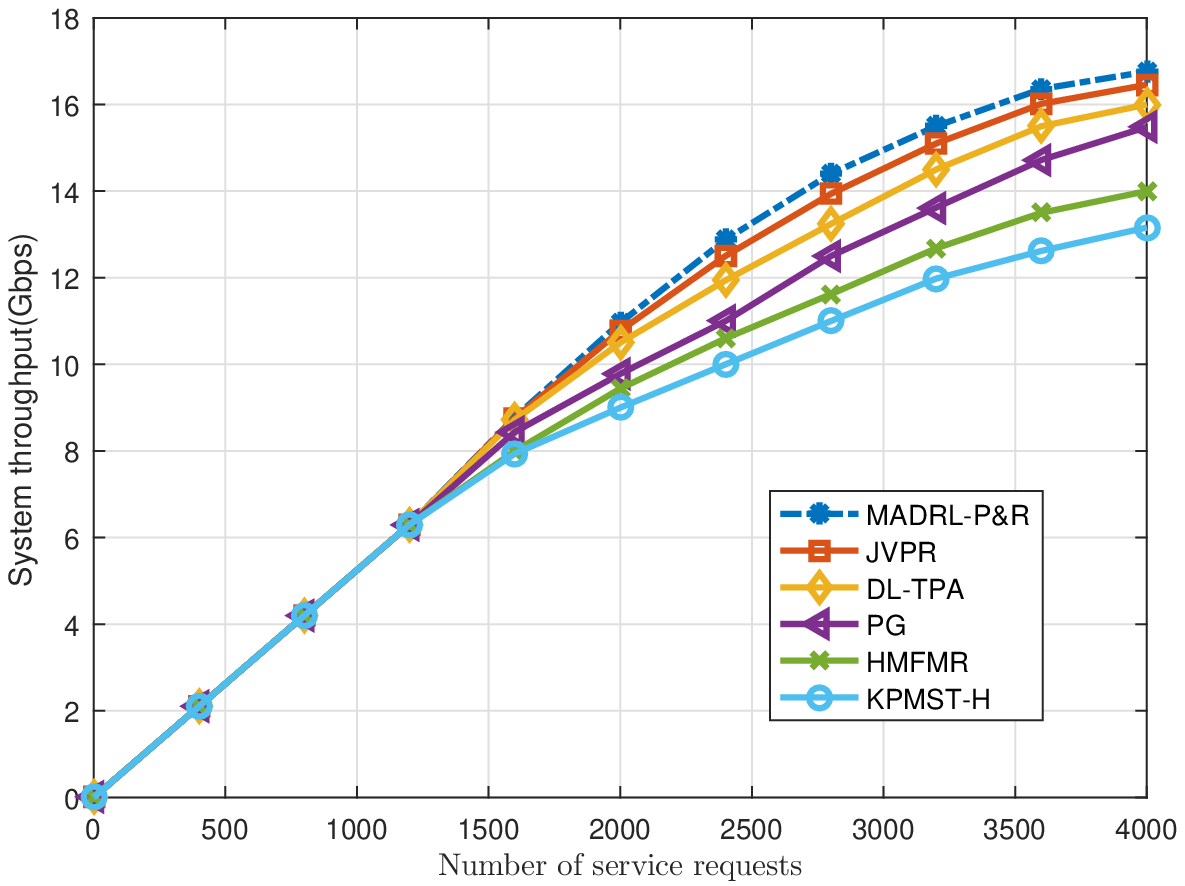}
  }
  \subfigure[$r_{i}^{\mathrm{b}}$=5.4 Mbps, $\overline{\varphi_{i}^{\mathrm{d}}}=0.5$]{
    \includegraphics[width=3.1in]{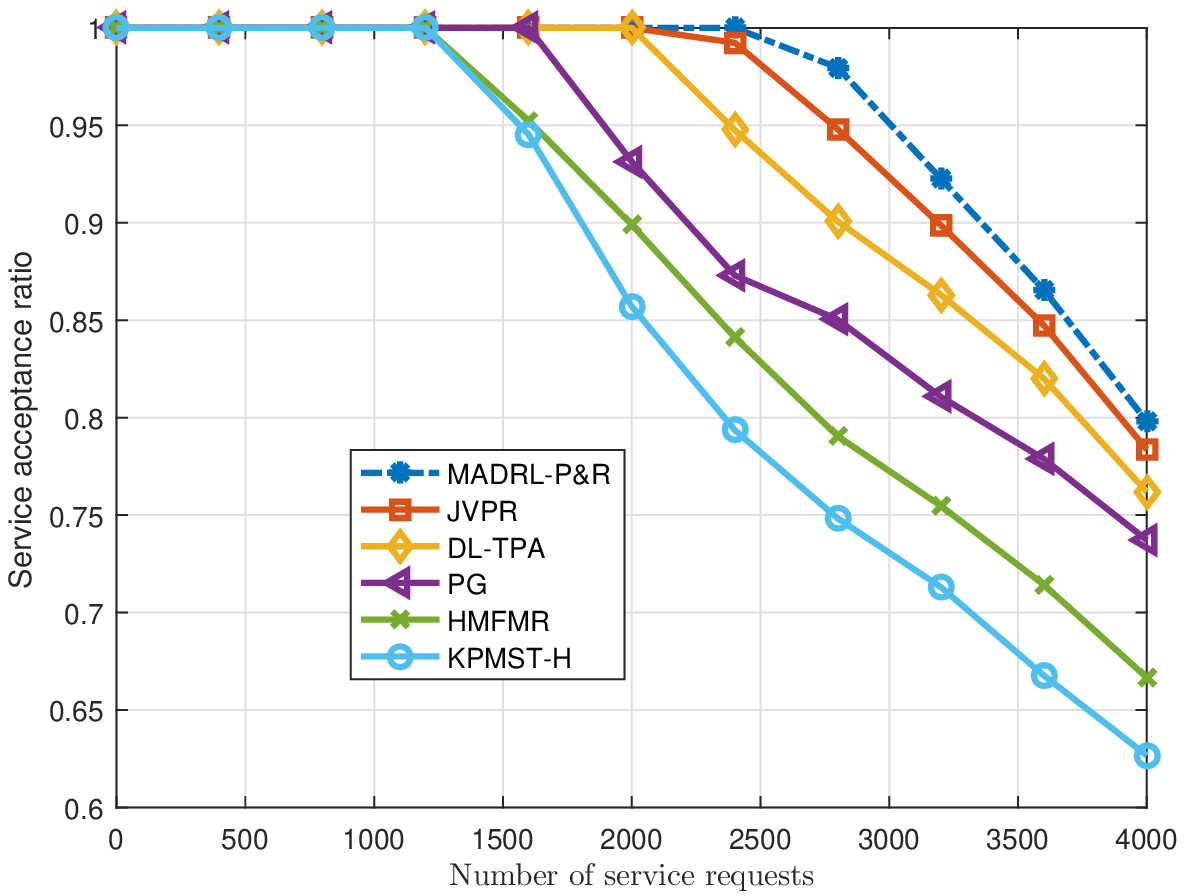}
  }
  \caption{TA2 network topology. (a): Comparison of system throughput; (b): comparison of service acceptance ratio.}
  \label{throughput_sar_TA2_R3}
\end{figure}
\begin{figure}[h]
  \centering
  \subfigure[$r_{i}^{\mathrm{b}}$=5.4 Mbps, $\overline{\varphi_{i}^{\mathrm{d}}}=0.5$]{
    \includegraphics[width=3.1in]{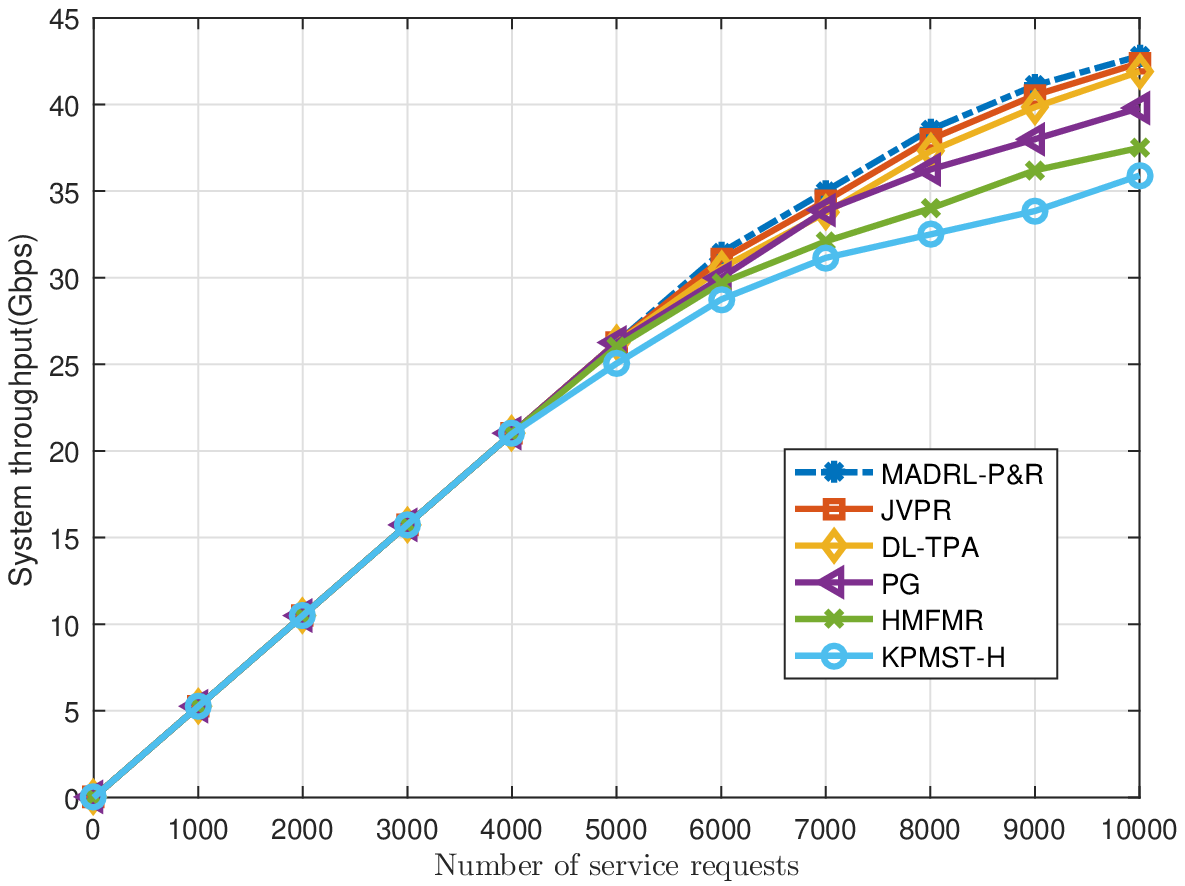}
  }
  \subfigure[$r_{i}^{\mathrm{b}}$=5.4 Mbps, $\overline{\varphi_{i}^{\mathrm{d}}}=0.5$]{
    \includegraphics[width=3.1in]{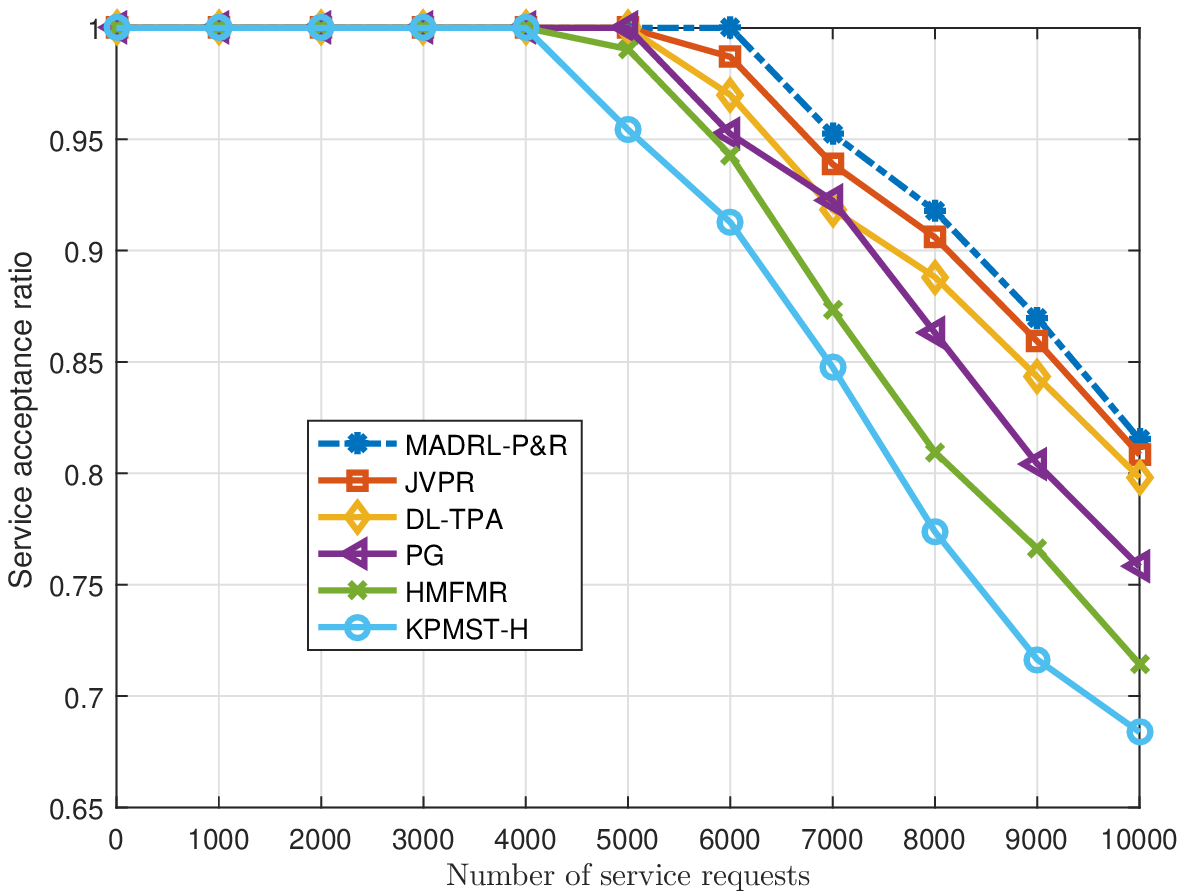}
  }
  \caption{SELF1 network topology. (a): Comparison of system throughput; (b): comparison of service acceptance ratio.}
  \label{throughput_sar_self1_R3}
\end{figure}
\begin{figure}[h]
  \centering
  \subfigure[$r_{i}^{\mathrm{b}}$=5.4 Mbps, $\overline{\varphi_{i}^{\mathrm{d}}}=0.5$]{
    \includegraphics[width=3.1in]{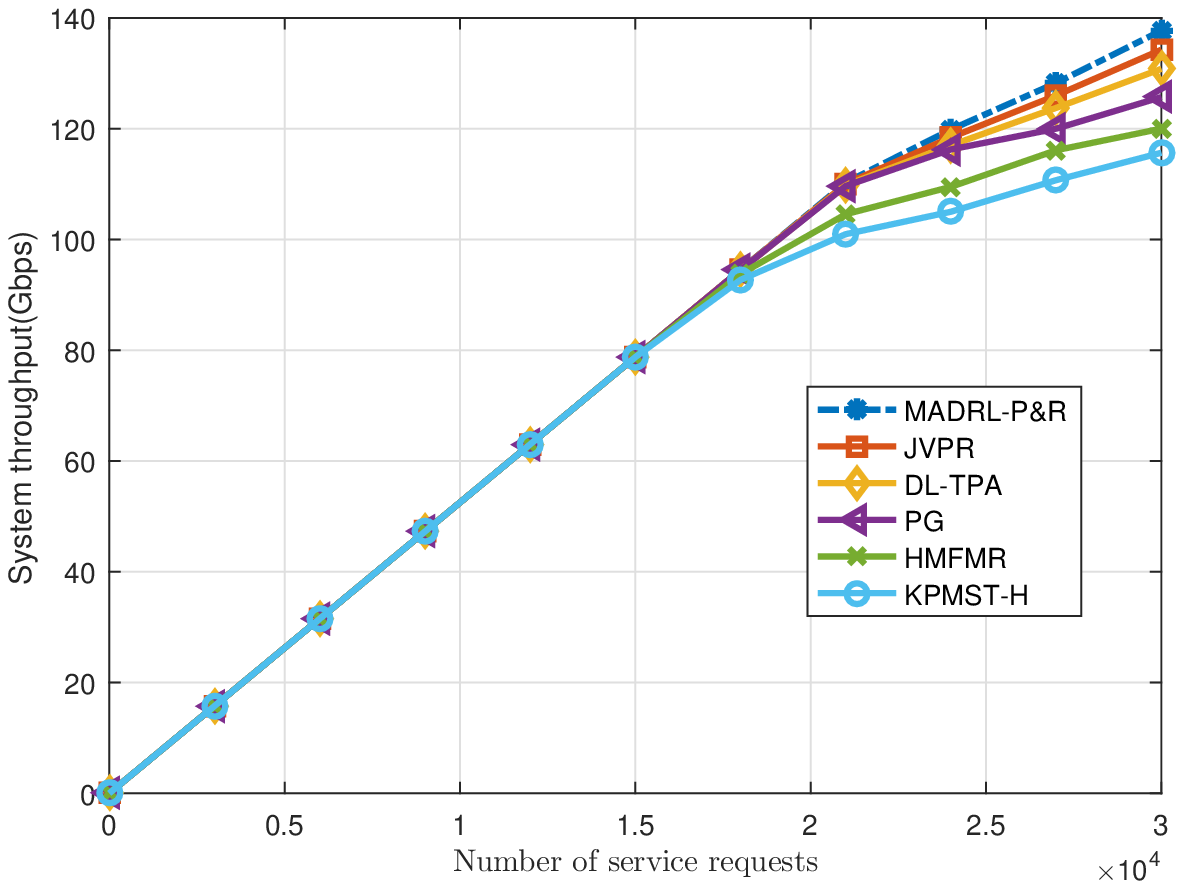}
  }
  \subfigure[$r_{i}^{\mathrm{b}}$=5.4 Mbps, $\overline{\varphi_{i}^{\mathrm{d}}}=0.5$]{
    \includegraphics[width=3.1in]{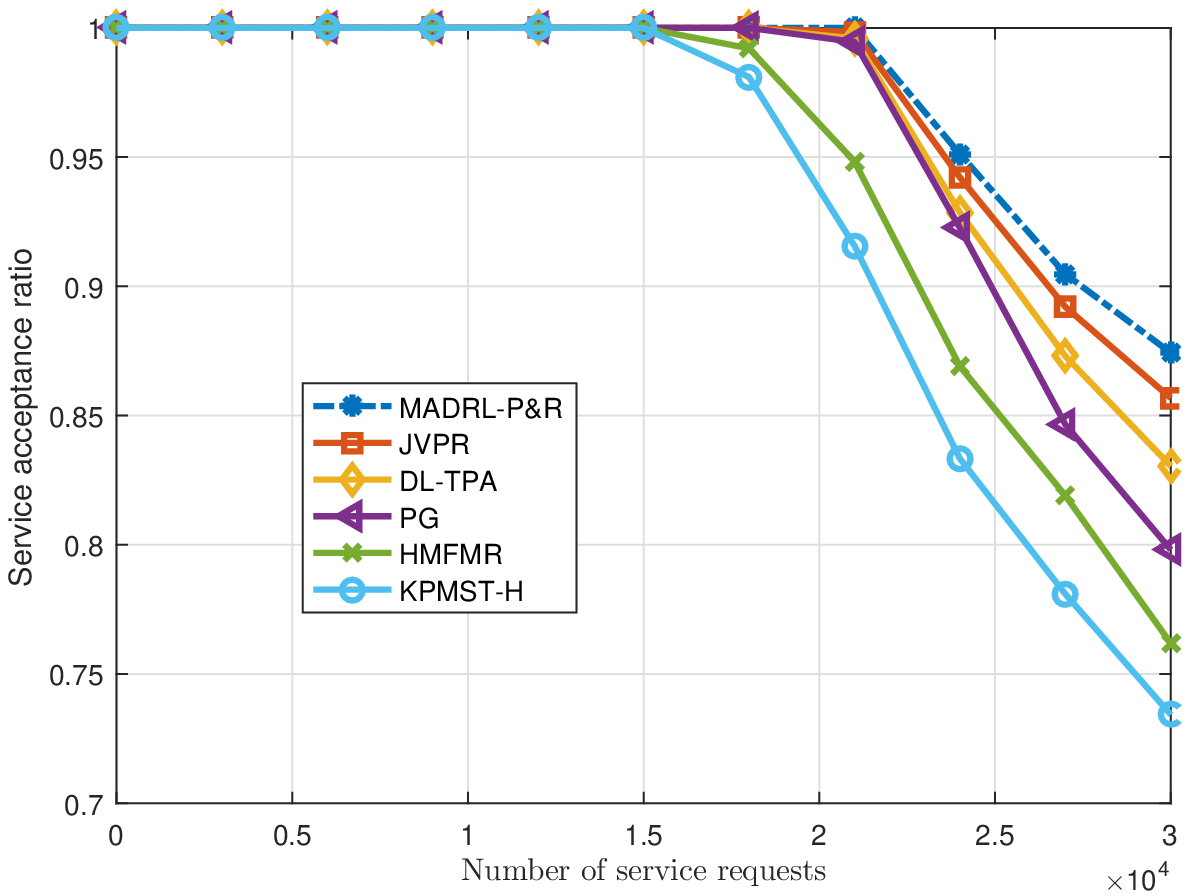}
  }
  \caption{SELF2 network topology. (a): Comparison of system throughput; (b): comparison of service acceptance ratio.}
  \label{throughput_sar_self2_R3}
\end{figure}
\begin{figure}[h]
\centering
\includegraphics[scale=0.6]{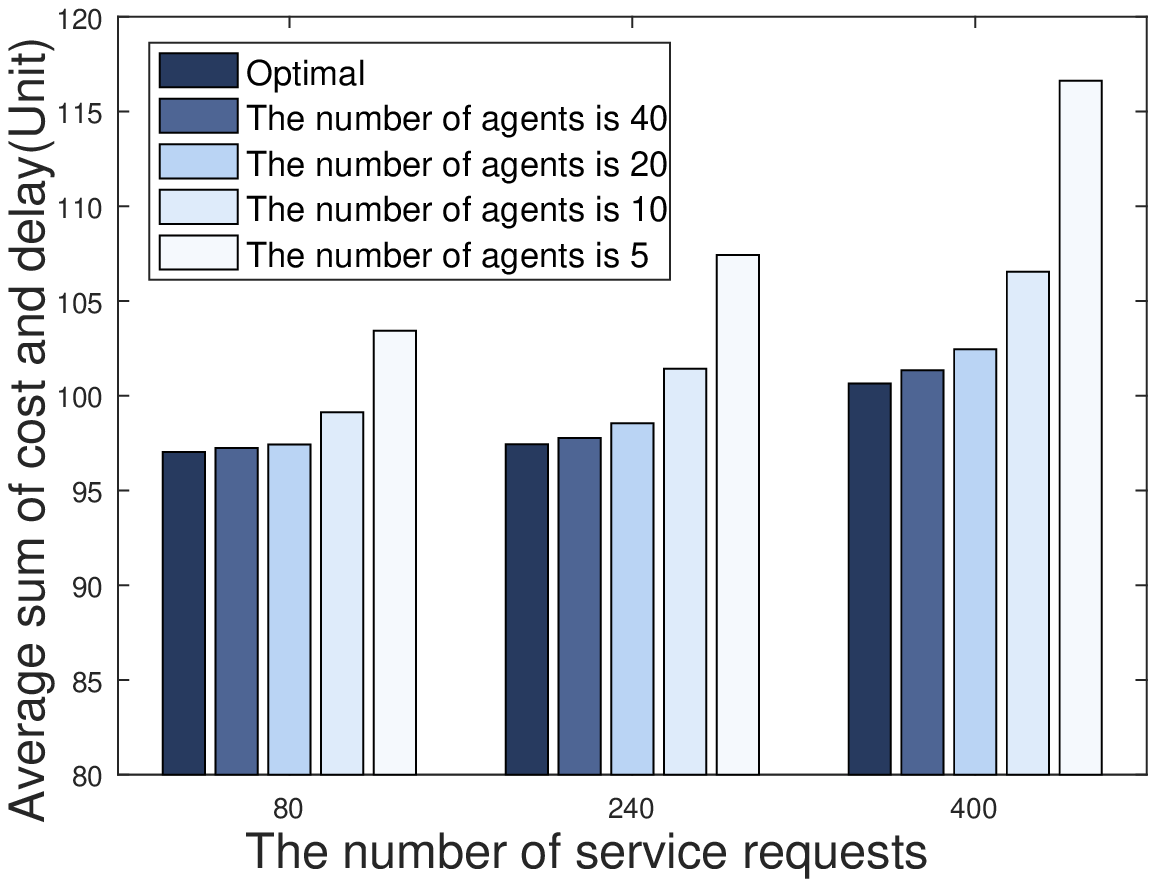}
\caption{The comparison of MADRL-P\&R algorithm based on different numbers of agents.}
\label{different_agents}
\end{figure}

Fig. \ref{throughput_sar}(a) compares changes in system throughput with the number of service requests. The system throughput increases linearly as the number increases when the number of service requests is small. The growth of system throughput is hindered when the number of service requests is large. This is because limited by the network capacity, the current resources are not enough to support all service requests simultaneously. In Fig. \ref{throughput_sar}(b), the service acceptance ratio means service requests successfully served (i.e., the P\&R policy can meet the requested service rate $r_{i}^{\mathrm{b}}$) accounting for the total service requests. As the number of requests increases, the service acceptance rate gets lower. MADRL-P\&R performs the best, indicating that the MADRL-P\&R can offer more reasonable placement and routing to more efficiently utilize network resources. This conclusion can also be further demonstrated in Figs. \ref{throughput_sar_TA2_R3}, \ref{throughput_sar_self1_R3} and \ref{throughput_sar_self2_R3}.
As analyzed in Section V-C, in the actual MADRL-P\&R implementation, the $M$ service requests are divided into smaller batches to perform deployment. Fig. \ref{different_agents} compares the average sum of cost and delay obtained by MADRL-P\&R under different numbers of agents. A result close to the optimal can be achieved when the number of agents is 20 and 40. Taking into account the limitation of running time, the experiments in this paper are all realized by MADRL-P\&R with 20 agents.
\begin{figure}[h]
  \centering
  \subfigure[]{
    \includegraphics[scale=0.32]{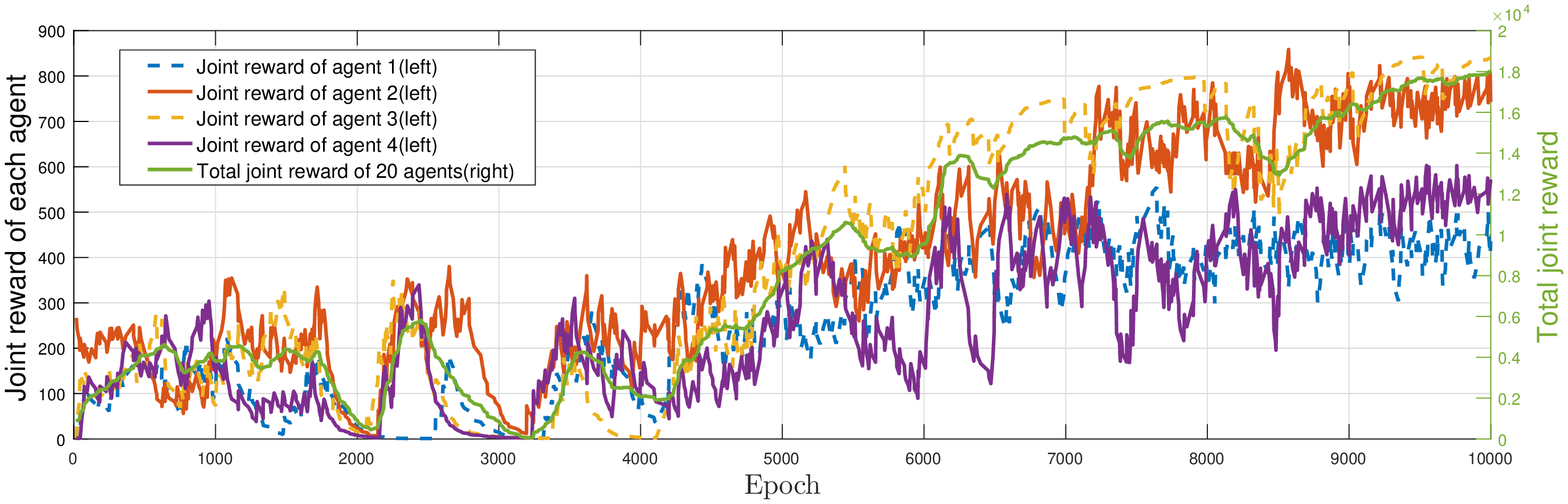}
  }
  \subfigure[]{
    \includegraphics[scale=0.32]{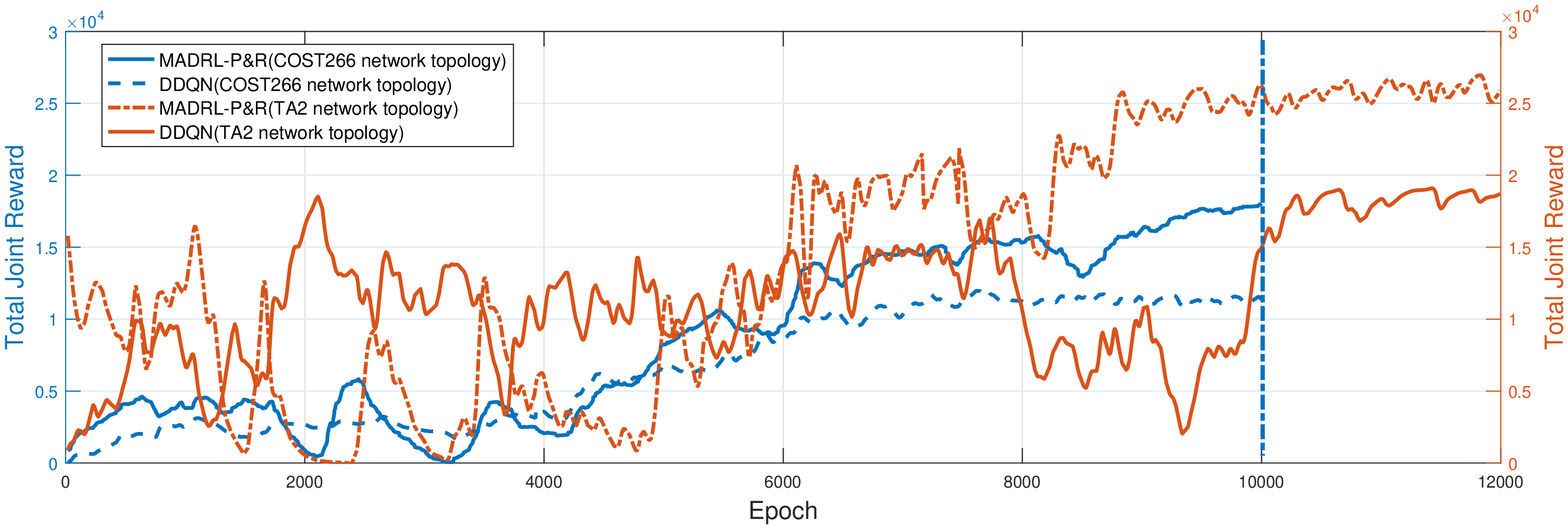}
  }
  \subfigure[]{
    \includegraphics[scale=0.32]{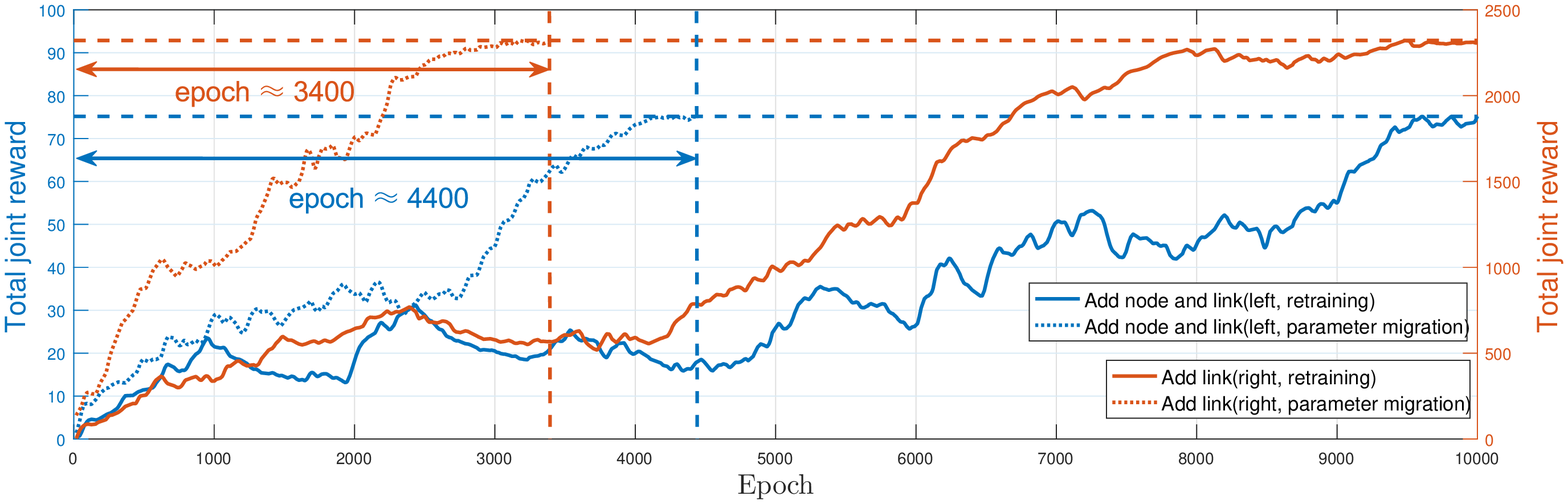}
  }
  \caption{(a): The comparison of joint reward of each agent and total joint reward of all agents; (b): The convergence comparison of our method with DDQN on COST266 and TA2 network topologies; and (c): The training process of parameter migration-based and retraining-based methods when the network topology changes.}
  \label{convergence_total}
\end{figure}
\subsection{Convergence of Different Agents and Influence of Topology Change}
Fig. \ref{convergence_total}(a) shows the training process of MADRL-P\&R model (due to space limitations, only four PA-agents are shown). All agents exhibit good convergence. Compared with the total joint reward, the joint reward of each agent changes more drastically. On the one hand, it is a result of the interaction between different agents. On the other hand, the sum of the joint rewards of the 20 agents weakens the impact of the change of a single agent.

The convergence comparison of our method with typical DRL algorithms: DDQN, on different network topologies, is shown in Fig. \ref{convergence_total}(b). Our proposed MADRL-P\&R method has higher performance, although the stability of DDQN is better than our method.

Fig. \ref{convergence_total}(c) shows the training process of parameter migration-based and retraining-based methods when the network topology changes. Our proposed parameter migration-based model-retraining method converges faster than the retraining-based method, especially when only links change. The convergence speed is improved by 66\%. This is because the parameter migration-based method inherits the capabilities of the previously trained model and only needs a little additional training to achieve convergence. Compared with only changing links (e.g., adding five links), changing both links and nodes (e.g., adding one node and four links) leads to more drastic network topology changes, and longer training time is required until convergence.
\section{Conclusion}
This paper addresses the minimization of cost and delay problem in joint VNF placement and routing by leveraging the MADRL technique, and developes a new MADRL-P\&R framework. Experiments confirmed that the proposed MADRL-P\&R approach is better than the existing alternative algorithms for the service cost and delay. Our MADRL-P\&R can achieve more reasonable resource allocation and hence improve system throughput. In addition, our approach is more sensitive to user differentiation, accommodating flexible services for individual user demands. Our parameter migration-based model-retraining method can effectively cope with network topology changes. When the network topology changes less, the new parameter migration-based model-retraining method can considerably accelerate the model training.
\bibliographystyle{IEEEtran}

\begin{IEEEbiography}[{\includegraphics[width=1in,height=1.25in,clip,keepaspectratio]{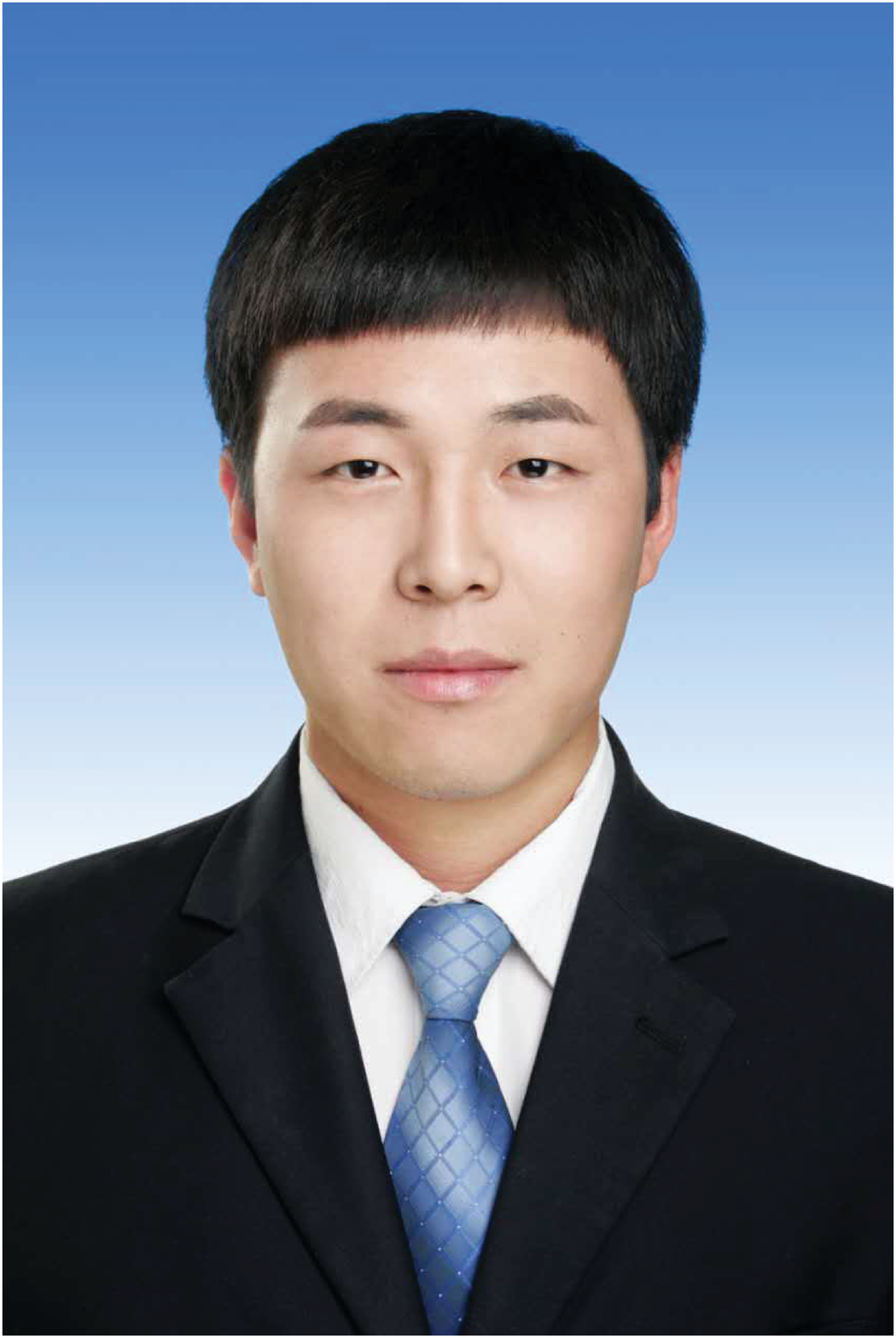}}]{Shaoyang Wang}
(S'18) received the B.E. degree in electronic information science and technology from Shandong University, China, in 2017. He is currently pursuing the Ph.D. degree with the School of Information and Communication Engineering, Beijing University of Posts and Telecommunications (BUPT), Beijing, China. His current research interests include non-orthogonal multiple access access, network function virtualization, and intelligent wireless resource management.
\end{IEEEbiography}
\begin{IEEEbiography}[{\includegraphics[width=1in,height=1.25in,clip,keepaspectratio]{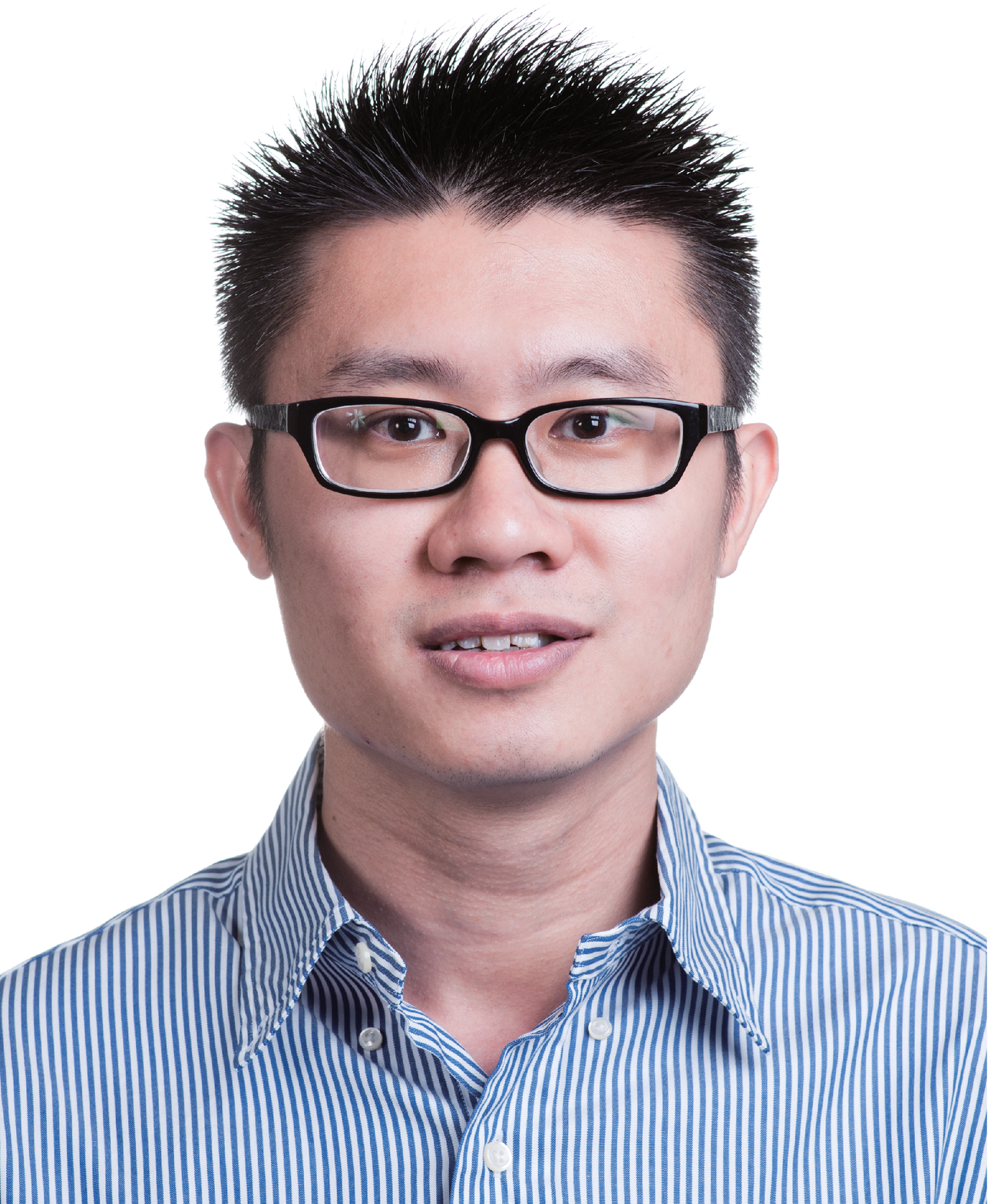}}]{Chau Yuen}
 (S'02-M'06-SM'12-F'21) received the B.Eng. and Ph.D. degrees from Nanyang Technological University (NTU), Singapore, in 2000 and 2004, respectively. He was a Post-Doctoral Fellow with Lucent Technologies Bell Labs, Murray Hill, in 2005. From 2006 to 2010, he was with the Institute for Infocomm Research (I2R), Singapore. Since 2010, he has been with the Singapore University of Technology and Design. Dr. Yuen was a recipient of the Lee Kuan Yew Gold Medal, the Institution of Electrical Engineers Book Prize, the Institute of Engineering of Singapore Gold Medal, the Merck Sharp and Dohme Gold Medal, and twice a recipient of the Hewlett Packard Prize. He received the IEEE Asia Pacific Outstanding Young Researcher Award in 2012 and IEEE VTS Singapore Chapter Outstanding Service Award on 2019. Currently, he serves as an Editor for the IEEE TRANSACTIONS ON VEHICULAR TECHNOLOGY, IEEE System Journal, and IEEE Transactions on Network Science and Engineering . He served as the guest editor for several special issues, including IEEE JOURNAL ON SELECTED AREAS IN COMMUNICATIONS, IEEE WIRELESS COMMUNICATIONS MAGAZINE, IEEE TRANSACTIONS ON COGNITIVE COMMUNICATIONS AND NETWORKING. He is a Distinguished Lecturer of IEEE Vehicular Technology Society.
\end{IEEEbiography}
\begin{IEEEbiography}[{\includegraphics[width=1in,height =1.25in,clip,keepaspectratio]{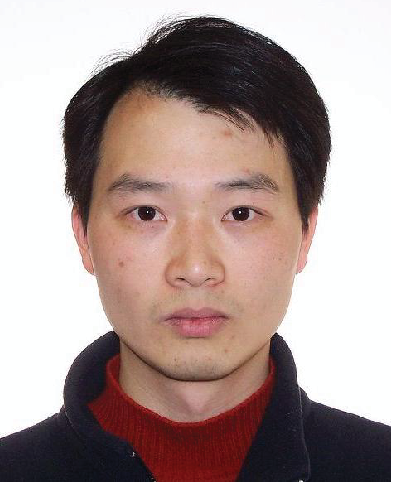}}]{Wei Ni} (M'09-SM'15) received the B.E. and Ph.D. degrees in Electronic Engineering from Fudan University, Shanghai, China, in 2000 and 2005, respectively. Currently, he is a Group Leader and Principal Research Scientist at CSIRO, Sydney, Australia, and an Adjunct Professor at the University of Technology Sydney and Honorary Professor at Macquarie University, Sydney. He was a Postdoctoral Research Fellow at Shanghai Jiaotong University from 2005 to 2008; Deputy Project Manager at the Bell Labs, Alcatel/Alcatel-Lucent from 2005 to 2008; and Senior Researcher at Devices R\&D, Nokia from 2008 to 2009. His research interests include signal processing, optimization, learning, and their applications to network efficiency and integrity.
\end{IEEEbiography}
\begin{IEEEbiography}[{\includegraphics[width=1in,height=1.25in,clip,keepaspectratio]{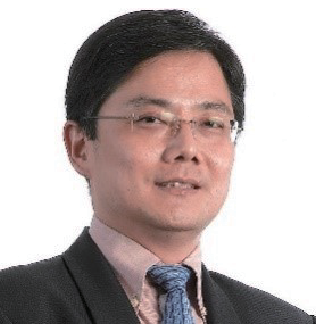}}]{Yong Liang GUAN} obtained his PhD from the Imperial College London, UK, and Bachelor of Engineering with first class honours from the National University of Singapore.  He is a Professor of Communication Engineering at the School of Electrical and Electronic Engineering, Nanyang Technological University (NTU), Singapore, where he leads the Continental-NTU Corporate Research Lab and the successful deployment of the campus-wide NTU-NXP V2X Test Bed.  His research interests broadly include coding and signal processing for communication systems and data storage systems.  He is an Editor for the IEEE Transactions on Vehicular Technology.  Currently he is also an Associate Vice President of NTU and a Distinguished Lecturer of the IEEE Vehicular Technology Society.
\end{IEEEbiography}
\begin{IEEEbiography}[{\includegraphics[width=1in,height=1.25in,clip,keepaspectratio]{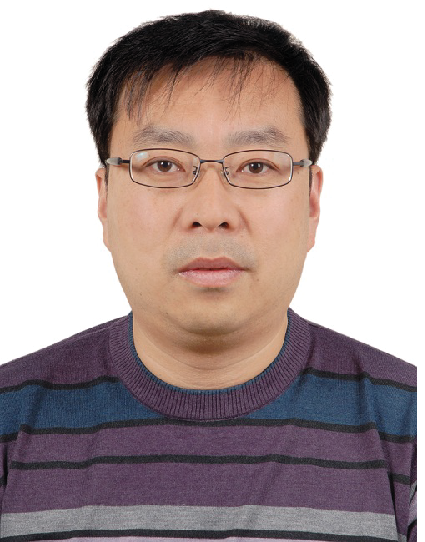}}]{Tiejun Lv}
(M'08-SM'12) received the M.S. and Ph.D. degrees in electronic engineering from the University of Electronic Science and Technology of China (UESTC), Chengdu, China, in 1997 and 2000, respectively. From January 2001 to January 2003, he was a Postdoctoral Fellow with Tsinghua University, Beijing, China. In 2005, he was promoted to a Full Professor with the School of Information and Communication Engineering, Beijing University of Posts and Telecommunications (BUPT). From September 2008 to March 2009, he was a Visiting Professor with the Department of Electrical Engineering, Stanford University, Stanford, CA, USA. He is the author of three books, more than 100 published IEEE journal papers and 200 conference papers on the physical layer of wireless mobile communications. His current research interests include signal processing, communications theory and networking. He was the recipient of the Program for New Century Excellent Talents in University Award from the Ministry of Education, China, in 2006. He received the Nature Science Award in the Ministry of Education of China for the hierarchical cooperative communication theory and technologies in 2015.
\end{IEEEbiography}
\end{document}